\documentclass[runningheads,a4paper]{llncs}
\usepackage{geometry}
\usepackage{graphicx}
\usepackage{wrapfig}
\usepackage{caption}
\usepackage{subfig}
\usepackage{todonotes}
\usepackage{amssymb}
\usepackage{amsfonts}
\usepackage[mathscr]{euscript}
\usepackage{mathtools}
\usepackage{sidecap}
\newcommand\blfootnote[1]{%
  \begingroup
  \renewcommand\thefootnote{}\footnote{#1}%
  \addtocounter{footnote}{-1}%
  \endgroup
}
\begin{document}
\title{
Evolution of Hierarchical Structure \& Reuse in iGEM Synthetic DNA Sequences
}
%
%
\author{Payam Siyari\inst{1} \and
Bistra Dilkina\inst{2} \and
Constantine Dovrolis\inst{1}}
\authorrunning{P. Siyari et al.}
%
\institute{Georgia Institute of Technology, Atlanta GA 30332, USA\\
\email{\{payamsiyari, constantine\}@gatech.edu}
\and
University of Southern California, Los Angeles CA 90007, USA
\\\email{dilkina@usc.edu}}
\maketitle              
\vspace{-5mm}
\begin{abstract}
Many complex systems, both in technology and nature, exhibit hierarchical modularity: smaller modules, each of them providing a certain function, are used within larger modules that perform more complex functions. 
Previously, we have proposed a modeling framework, referred to as Evo-Lexis \cite{siyari2018}, that provides insight to some fundamental questions about evolving hierarchical systems.

The predictions of the Evo-Lexis model should be tested using real data from evolving systems in which the outputs can be well represented by sequences. In this paper, we investigate the time series of iGEM synthetic DNA dataset sequences, and whether the resulting iGEM hierarchies exhibit the qualitative properties predicted by the Evo-Lexis framework. Contrary to Evo-Lexis, in iGEM the amount of reuse decreases during the timeline of the dataset. Although this results in development of less cost-efficient and less deep Lexis-DAGs, the dataset exhibits a bias in reusing specific nodes more often than others. This results in the Lexis-DAGs to take the shape of an hourglass with relatively high H-score values and stable set of core nodes. Despite the reuse bias and stability of the core set, the dataset presents a high amount of diversity among the targets which is in line with modeling of Evo-Lexis.

\end{abstract}
\blfootnote{
\scriptsize This research is supported by DARPA’s Lifelong Learning Machines (L2M) program, under Cooperative Agreement HR0011-18-2-0019, and by the National Science Foundation under Grant No. 1319549.
}

\section{Introduction}
Hierarchically modular designs enhance evolvability in natural systems \cite{clune2016,hie1,sabrin2016}, make the maintenance easier in technological systems, and provide agility and better abstraction of the system design \cite{hie2,myers}.

In prior work in \cite{siyari2018}, we present \emph{Evo-Lexis}, a modeling framework for the emergence and evolution of hierarchical structure in complex modular systems. There are many hypotheses in the literature regarding the factors that contribute to either the hierarchy or modularity properties.
Local resource constraints in social networks and ecosystems \cite{Miller2008}, modularly varying goals \cite{clune2012,kashtan2007,kashtan2005}, selection for more robust phenotypes \cite{clune-robust-pheno2,clune-robust-pheno1}, and selection for lower connection costs in a network \cite{clune2016} are some of the mechanisms that have been previously explored and shown to lead to hierarchically modular systems. 
 The main hypothesis that Evo-Lexis follows is along the lines of \cite{clune2016},  which assumes that systems in both nature and technology care to minimize the cost of their interconnections or dependencies between modules.
 We also studied the hourglass effect via Evo-Lexis. Informally, an hourglass architecture means that the system of interest produces many outputs from many inputs through a relatively small number of highly central intermediate modules, referred to as the ``waist'' of the hourglass.
It has been observed that hierarchically modular systems often exhibit the architecture of an  hourglass; for reference, in fields like computer networking \cite{akhshabi2011}, neural networks \cite{hourglass-dnn,alon2015}, embryogenesis \cite{hourglass-embryo}, metabolism \cite{hourglass-metabolism2,hourglass-metabolism},  and many others \cite{sabrin2016,bowtiebuilder}, this phenomena is observed.
A comprehensive survey of the literature on hierarchical systems evolution, and the hourglass effect is presented in \cite{sabrin2016}. 

The motivation for this paper is that the Evo-Lexis model is quite general and abstract,  and it does not attempt to capture any domain-specific aspects of biological or technological evolution. As such, it makes several assumptions that can be criticized for being unrealistic, such as the fact that all targets have the same length, or their length stays constant, or the fitness of a sequence is strictly based on its hierarchical cost.
We believe that such abstract modeling is still valuable because it can provide insights into the qualitative properties of the resulting hierarchies under different target generation models.
However, we also believe that the predictions of the Evo-Lexis model should be tested using real data from evolving systems in which the outputs can be well represented by sequences. One such system is the iGEM synthetic DNA dataset \cite{igem}. The target DNA sequences in the iGEM dataset are built from standard ``BioBrick parts'' (more elementary  DNA sequences) that collectively form a library of synthetic DNA sequences. These sequences are submitted to the registry of standard biological parts in the annual iGEM competition. Previous research in \cite{dnaSynthPaper,siyari2016a} has provided some evidence that these synthetic DNA sequences are designed by reusing existing components, and as such, it has a hierarchical organization. In this paper, we investigate how to apply the Evo-Lexis framework in the time series of iGEM sequences, and whether the resulting iGEM hierarchies exhibit the same qualitative properties we observed in \cite{siyari2018} which was solely based on abstract target generation models.
We ask the following questions in this paper:
\begin{enumerate}
\item 
How can we analyze the iGEM dataset using the evolutionary framework of Evo-Lexis? How are the batches of targets formed? What properties of the iGEM batches are different than Evo-Lexis's setting?
\item
When formed incrementally over the iGEM dataset, which are the architectural properties of Lexis-DAGs, and why? 

\end{enumerate}
\section{Preliminaries}
\label{sec:lexis-prelim}
To develop \emph{Evo-Lexis}, we extend the previously proposed optimization framework \emph{Lexis} in \cite{siyari2016a}. Lexis models the most elementary modules of the system as symbols (``sources'') and the modules at the highest level of the hierarchy as sequences of those symbols (``targets''). \emph{Evo-Lexis} is a dynamic or evolving version of Lexis, in the sense that the set of targets changes over time through additions (births) and removals (deaths) of targets. \emph{Evo-Lexis} computes an (approximate) minimum-cost adjustment of a given hierarchy when the set of targets changes over time (a process we refer to as ``incremental design'').

\subsection{Lexis Optimization}
\begin{SCfigure}
\centering
\subfloat[]{
\includegraphics[trim = 1.3cm 1.35cm 1.40cm 1.39cm,clip,scale=0.3]{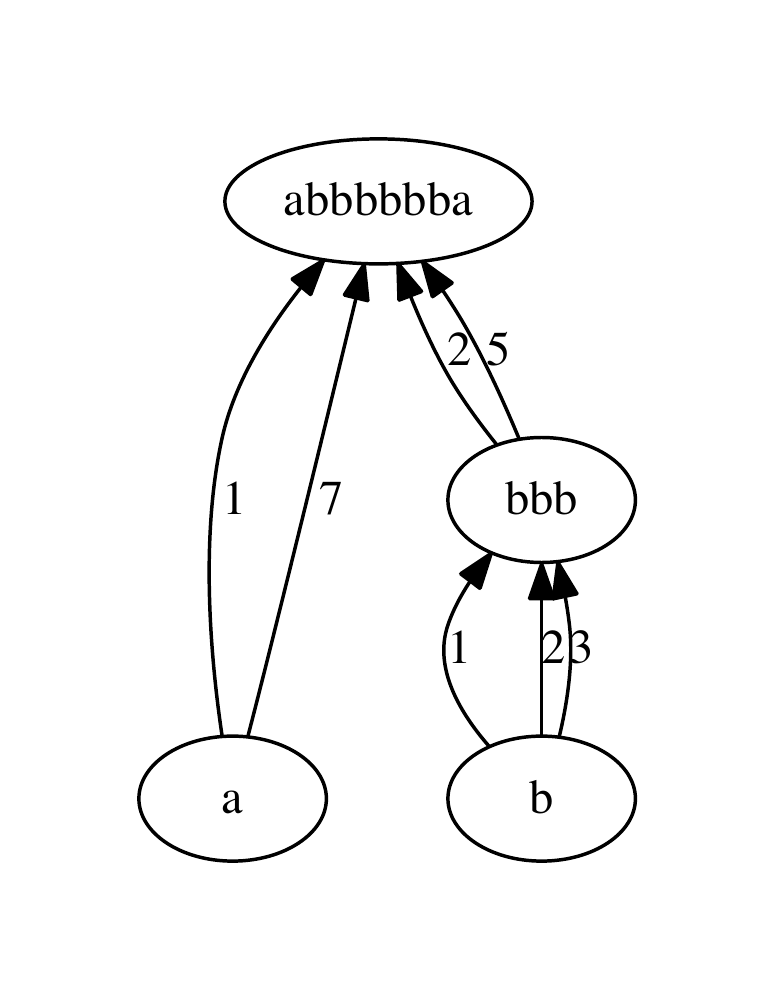}
}
\hspace{.25cm}
  \subfloat[]{
  \includegraphics[trim = 1.4cm 1.35cm 1.35cm 1.39cm,clip,scale=0.3]{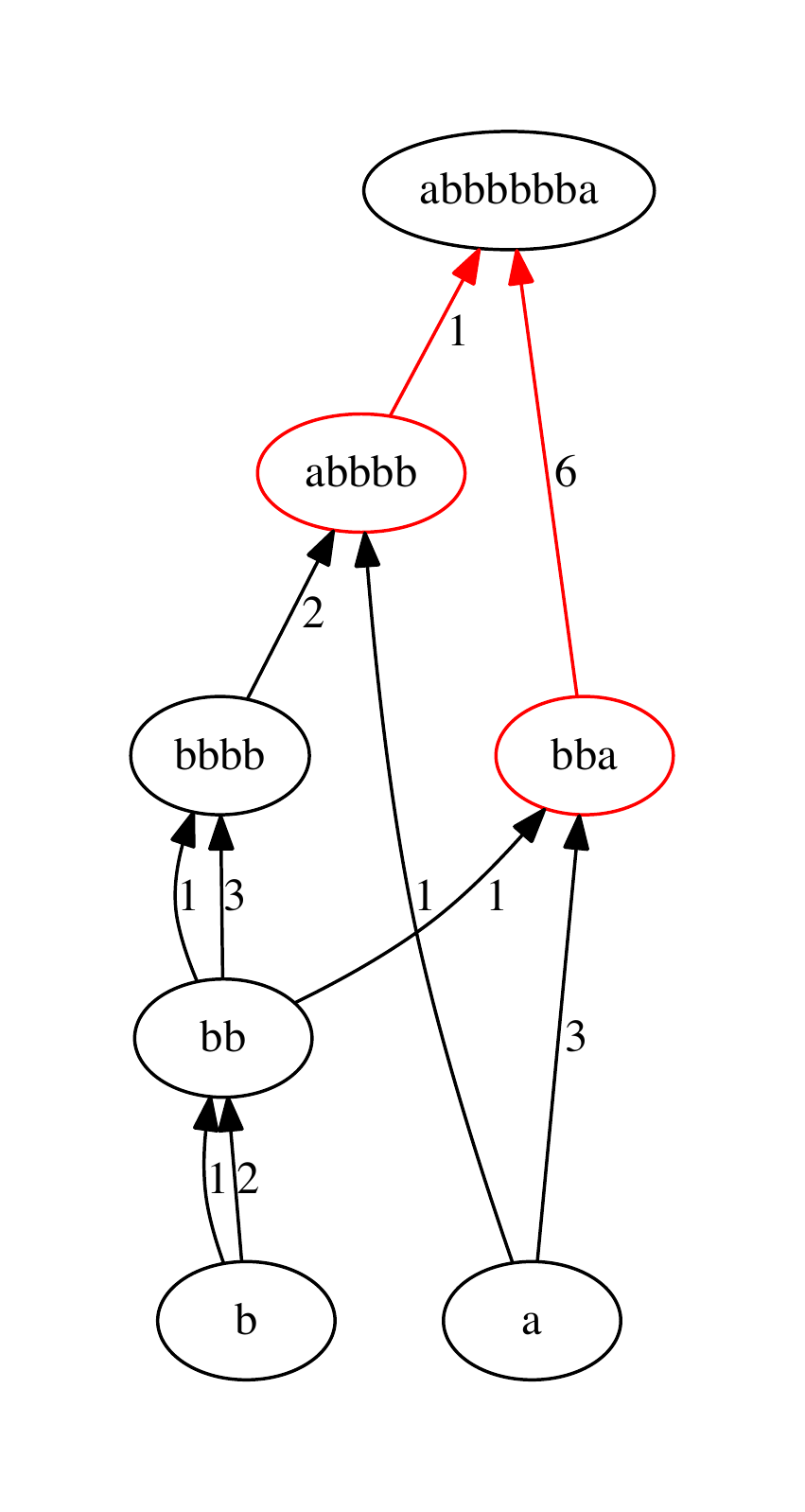}
  }
    \subfloat[]{
    \includegraphics[trim = 1.3cm 1.35cm 1.35cm 1.39cm,clip,scale=0.3]{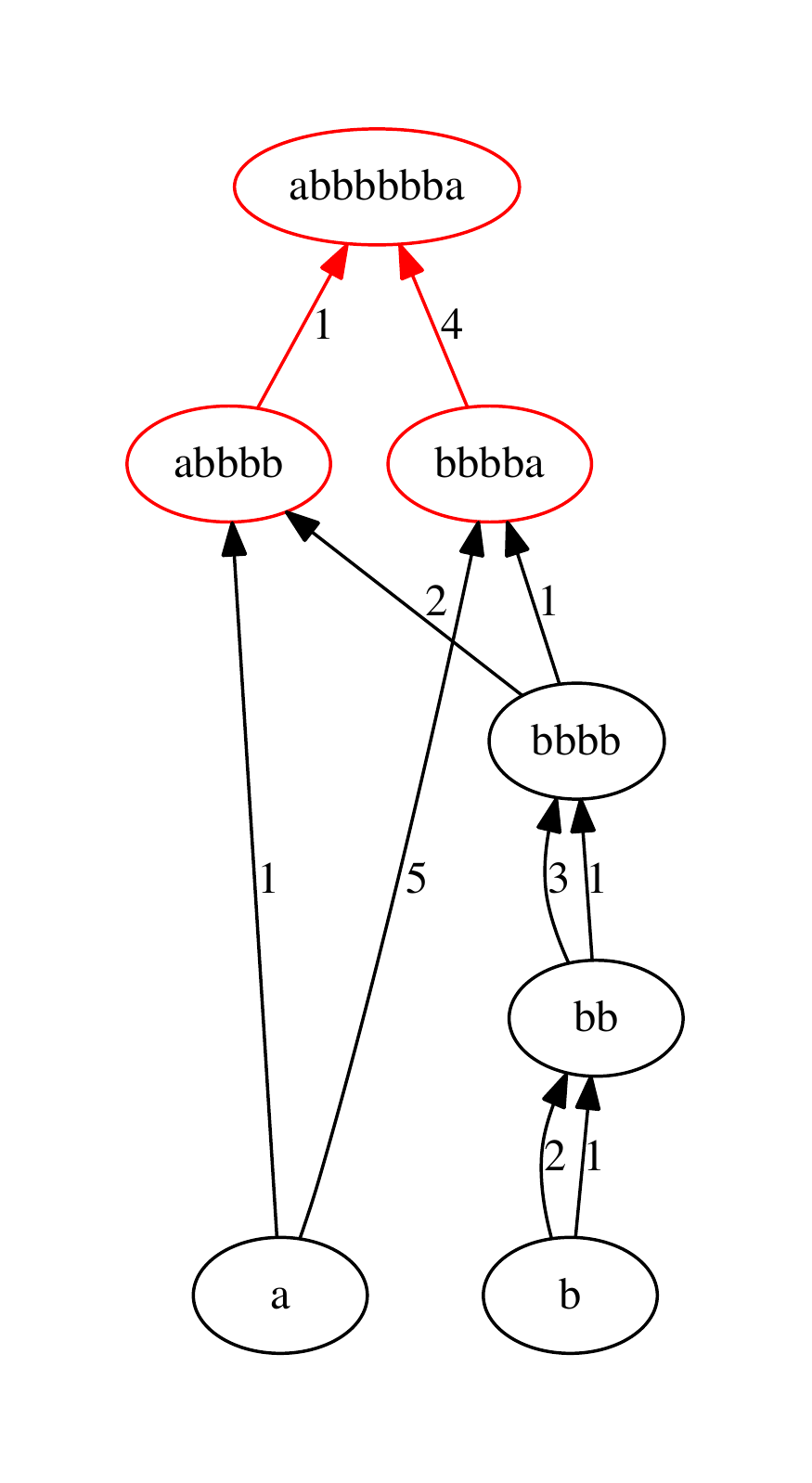}
    }
\caption{\small Illustration of the Lexis-DAG  for a single target $T=\{abbbbbba\}$ and sources $S=\{a,b\}$. Edge-labels indicate the occurrence indices: {\bf (a)} A valid Lexis-DAG having both minimum number of concatenations and edges. {\bf (b)} An invalid Lexis-DAG: two intermediate nodes are re-used only once. {\bf (c)} An invalid Lexis-DAG: the top-layer string is not equal to the concatenation of its two in-neighbors (best viewed in color).}
\label{fig:dagDef}
\end{SCfigure}

Given an alphabet $S$ and a set of ``target'' strings $T$ over the alphabet $S$, we need to construct a Lexis-DAG.
A Lexis-DAG $D$ is a directed acyclic graph $D(V,E)$, where $V$ is the set of nodes and $E$ the set of edges, that satisfies the following three constraints:\footnote{To simplify the notation, even though $D$ is 
a function of $S$ and $T$, we do not denote it as such.} {\bf a)} Each node $v \in V$ in a Lexis-DAG represents a string $\mathcal{S}(v)$
of characters from the alphabet $S$.
The nodes $V_S$ that represent characters of $S$ are referred to as \emph{sources}, and they have zero in-degree.
The nodes $V_T$ that represent target strings $T=\{t_1, t_2, \dots, t_m\}$ are referred to as {\em targets}, and they
have zero out-degree.
$V$ also includes a set of {\em intermediate nodes} $V_M$, which represent substrings that appear in the targets $T$.  
So, $V=V_S\cup V_M\cup V_T$. {\bf b)} 
Each node in $V_M\cup V_T$ of a Lexis-DAG represents a string that is the concatenation 
of two or more substrings, specified by the incoming edges from other nodes to that node. 
Note that there may be more than one edge from node $u$ to node $v$. 
{\bf c)} A Lexis-DAG should only include intermediate nodes that have an out-degree of at least two,
$\forall v\in V_M, d_{out}(v) \geq 2$ for a more parsimonious hierarchical representation. 
Fig.~\ref{fig:dagDef} illustrates the concepts introduced here.
\subsubsection{The Lexis Optimization Problem}
The {\em Lexis} optimization problem is to construct a minimum-cost Lexis-DAG for the 
given alphabet $S$ and target strings $T$.
In other words, the problem is to determine the set of intermediate nodes $V_M$ and all
required edges $E$ so that the corresponding Lexis-DAG $D$ is optimal in terms of a given cost
function $C(D)$. This problem can be formulated as follows: 

\begin{equation}
\label{eq:Lexis}
\begin{aligned}
\text{~}&min_{(E,V_{M})}\text{~}C(D)\\
\text{~}&s.t.\text{~} D=(V,E)\text{ is a Lexis-DAG for $S$ and $T$}\\
\text{~}&\text{where~} C(D)=\mathcal{E}(D) = \sum_{v\in V} d_{in}(v) = |E|
\end{aligned}
\end{equation}

A natural cost function, as investigated in previous work \cite{siyari2016a}, is the number of edges in the Lexis-DAG. 
The {\em edge cost} to construct a node $v \in V$ is defined as the number of incoming edges required to construct $\mathcal{S}(v)$ from its in-neighbors, which is equal to $d_{in}(v)$. The edge cost of source nodes is obviously zero. The edge cost $\mathcal{E}(D)$ of Lexis-DAG $D$ is defined as the edge cost of all nodes, which is equal to the number of edges in $D$.
With edge cost, the problem in Eq. \eqref{eq:Lexis} is NP-Hard \cite{siyari2016a}. This problem is similar to the \emph{Smallest Grammar Problem} (SGP) \cite{sgp} and in fact its NP-Hardness is shown by a reduction from SGP \cite{siyari2016a}.

We solve the Lexis optimization problem in Eq. \eqref{eq:Lexis} with a greedy heuristic, called \textsc{G-Lexis} \cite{siyari2016a}. 
\textsc{G-Lexis} starts with the trivial flat Lexis-DAG, and at each iteration it chooses the substring $\xi$  that maximally reduces the edge cost, when it is added as a new intermediate node to the Lexis-DAG and the corresponding edges are rewired by its addition.

\subsubsection{Path-Centrality and the Core of a Lexis-DAG}
After constructing a Lexis-DAG, an important question is to rank the constructed intermediate
nodes in terms of significance or {\em centrality.} 
More formally, let $P_D(v)$ be the number of source-to-target paths that traverse node $v \in V_M$;
we refer to $P_D(v)$ as the {\em path centrality} of intermediate node $v$.
Path centrality can be computed as: $P(v) = P_S(v) \, P_T(v)$
where $P_S(v)$  is the number of paths from any source to $v$, 
and $P_T(v)$ is the number of paths from $v$ to any target. \footnote{A similar metric, called \emph{stress centrality} of a vertex, is studied in \cite{sdm-cent}.} 

An important follow-up question is to identify the {\em core} of a Lexis-DAG, i.e., a set 
of intermediate nodes that represent, as a whole, the most important substrings in that Lexis-DAG. 
Intuitively, we expect that the core should include nodes of high path centrality, and that almost 
all source-to-target dependency chains of the Lexis-DAG should traverse at least one of these core nodes. More formally, suppose $K$ is a set of intermediate nodes 
and $\mathscr{P}^-(K)$ is the set of source-to-target paths after we remove the nodes in $K$ from $D$. 
The core of $D$ is defined as the minimum-cardinality set of intermediate nodes $Core(\tau)=\hat{K}$ such that
the fraction of remaining source-to-target paths after the removal of $\hat{K}$ is at most $\tau$:\footnote{To simplify notation, we do not denote the core set as function of $D$.}
\begin{equation}
\label{eq:coreid}
\begin{aligned}
\hat{K}=arg min_{\text{~}K \subseteq V_M}\text{~~~}&|K|\\
s.t.\text{~}&|\mathscr{P}^-(K)| \leq \tau \, |\mathscr{P}^-(\varnothing)| 
\end{aligned}
\end{equation}
where $|\mathscr{P}^-(\varnothing)|$ is the number of source-to-target paths in the original Lexis-DAG,
without removing any nodes.
We solve the core identification problem with a greedy algorithm referred to as \textsc{G-Core} \cite{siyari2016a}. 
This algorithm adds in each iteration the node with the highest path-centrality value to the core set, 
updates the Lexis-DAG by removing that node and its edges, and recomputes the path centralities of the remaining nodes before
the next iteration. 

\subsubsection{Hourglass score}
Intuitively, a Lexis-DAG exhibits the hourglass effect if it has a small core. 
We use a metric, named as Hourglass
Score, or \emph{H-Score}, in our study for measuring the ``hourglass-ness'' of a network. This metric was originally presented in \cite{sabrin2016}.
To calculate the H-score, we create a flat Lexis-DAG $D_f$ containing the same targets as the original Lexis-DAG $D$. Note that $D_f$ preserves the source-target dependencies of $D$: each target in $D_f$ is constructed based on the same set of sources as in $D$. However, the dependency paths in $D_f$ are direct, without forming any intermediate modules that could be reused across different targets. So, by construction, the flat Lexis-DAG $D_f$ cannot have a non-trivial core since it does not have any intermediate nodes. We define the H-score as follows: $H_D(\tau)=1- \frac{|Core(\tau)|}{|Core_f(\tau)|}$
where $Core(\tau)$ and $Core_f(\tau)$ are the core sets of $D$ and $D_f$ for a given threshold $\tau$, respectively. Since that $Core_f$ can include a combination of sources and targets, it would never be larger than either the set of sources or targets, i.e., $|Core_f(\tau)| \leq min\{|S|,|T|\}$. Thus, $0 \leq H(\tau) \leq 1$. The H-score of $D$ is approximately one if the core size of the original Lexis-DAG is negligible compared to the the core size of the corresponding flat Lexis-DAG.

\subsection{Evo-Lexis Framework and Key Results}
\label{evolexis-evo}

\begin{figure}[h!]
\center
\includegraphics[trim = 0cm 0cm 0cm 0cm,clip,width=.8\textwidth]{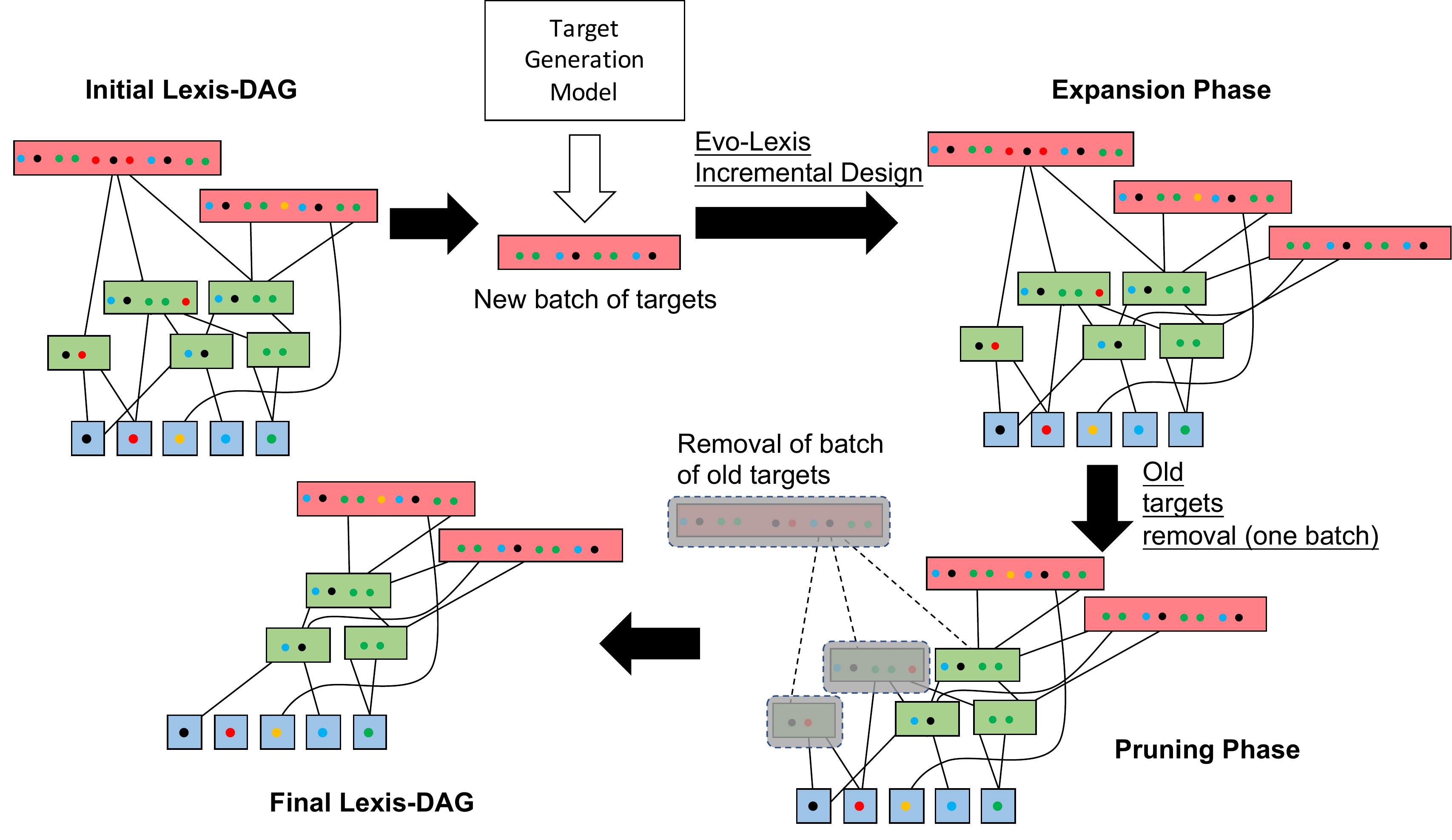}
\caption{A diagram of the Evo-Lexis framework.
}
\label{fig:evo-model}
\end{figure}

The Evo-Lexis framework includes a number of components that are described below. A general illustration of the framework is shown in Fig. \ref{fig:evo-model}.
In every iteration, the following steps are performed: {\bf (1)} A batch of new targets is generated via a target generation model. {\bf (2)} In the ``expansion phase'', the new targets are added incrementally to the current Lexis-DAG by minimizing the marginal cost of adding every new target to the existing hierarchy. We refer to this \emph{incremental design} algorithm as \textsc{Inc-Lexis}, and it is described in detail \cite{siyari2018}. {\bf (3)} If the number of targets that are present in the system has reached a steady-state threshold, we also remove the batch of oldest targets from the Lexis-DAG. 

\begin{figure}[h!]
\centering
\includegraphics[trim = 7.25cm 0cm 7.25cm 14.5cm,clip,width=.8\textwidth]{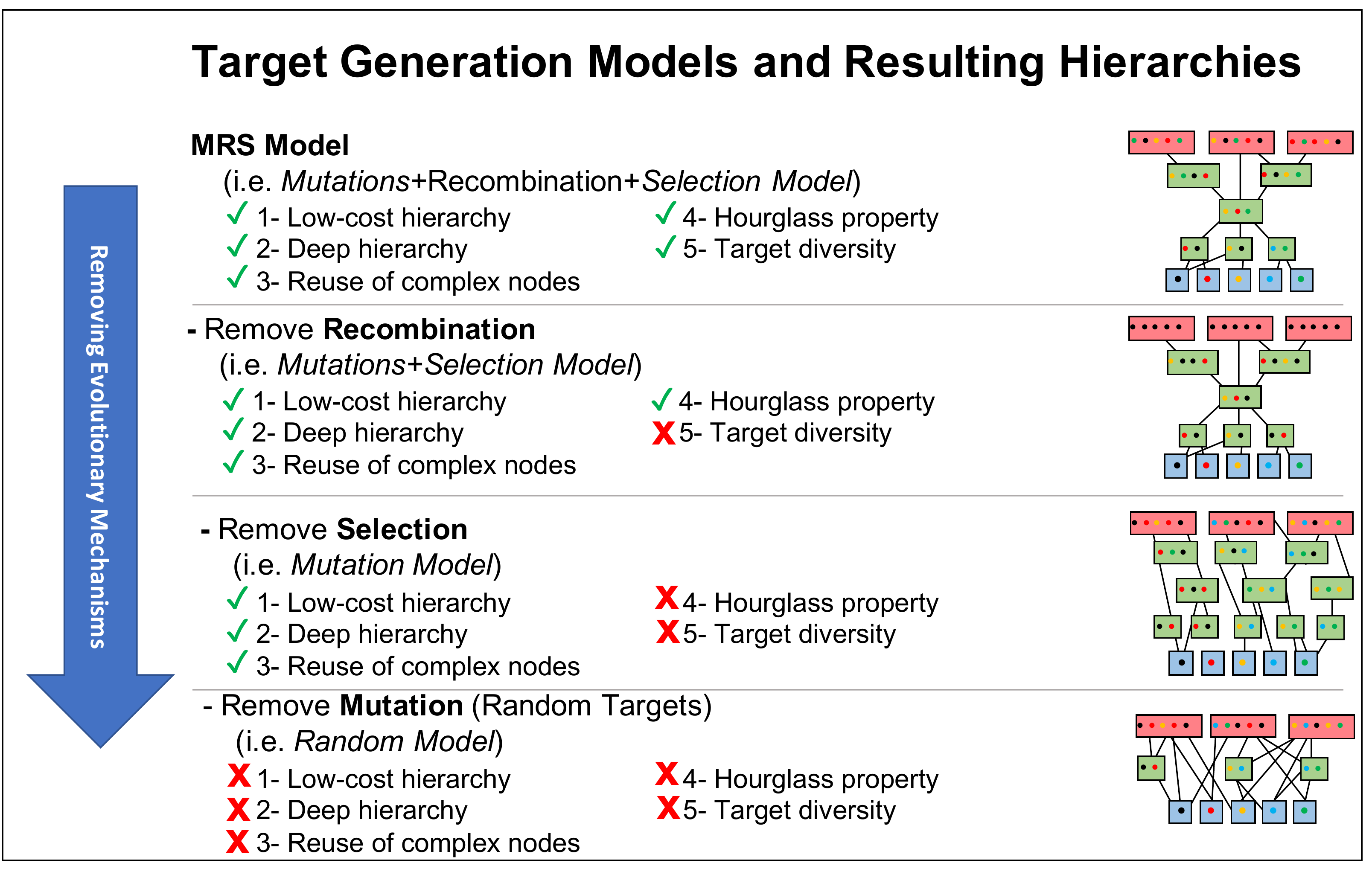}
\caption{{Overview of results from Evo-Lexis.}}
\label{fig:intro}
\end{figure}
In general, a system interacts with its environment in a bidirectional manner: the environment imposes various constraints on the system and the system also affects its environment. To capture this co-evolutionary setting in \emph{Evo-Lexis}, we study how changes in the set of targets affect the resulting hierarchy but also how the current hierarchy affects the selection of new targets (i.e. whether a new candidate target is selected or not depends on its fitness or cost -- and that depends on how easily that target can be supported by the given hierarchy). By incorporating well-known evolutionary mechanisms, such as tinkering (mutation), recombination, and selection, \emph{Evo-Lexis} can capture such co-evolutionary dynamics between the generation of new targets and the hierarchy that supports them. Fig. \ref{fig:intro} is an overview of the following key results from the Evo-Lexis model:
{\bf i)}
\emph{Tinkering/mutation} in the target generation process is found to be a strong initial force for the emergence of low-cost and deep hierarchies. 
{\bf ii)}
\emph{Selection} is found to enhance the emergence of more complex intermediate modules in optimized hierarchies.
The bias towards reuse of complex modules results in an hourglass architecture in which almost all source-to-target dependency paths traverse a small set of intermediate modules. 
{\bf iii)}
The addition of \emph{recombination} in the target generation process is essential in providing target diversity in optimized hierarchies. 

\section{iGEM Dataset}
\label{sec:dataset}
\subsection{Preliminaries}

The International Genetically Engineered Machine (iGEM) is an annual worldwide synthetic biology competition. The competition is between students from diverse backgrounds including biology, chemistry, physics, engineering, and computer science to construct synthetic DNA structures with novel functionalities.

Every year at the beginning of the summer, there is a ``Distribution Kit'' handed to teams which includes interchangeable parts (so called ``BioBricks'')  from the Registry of Standard Biological Parts comprising various genetic components such as promoters, terminators, reporter elements, and plasmid backbones. Then, the teams try to use these parts and the new standardized parts of their own in order to build biological systems. 
The teams can build on previous projects or create completely new parts. At the end of the summer, all teams add their new BioBricks to the registry for further possible reuse in next years. 

The iGEM Registry (i.e., the dataset we are working with) includes a set of standard biological parts. A [biological] part is a DNA sequence which encodes a biological function, e.g., a promoter or protein coding sequence. These biological parts are standardized to be easily assembled together and reused with other standardized parts in the registry. A ``basic part'' is a functional unit of a synthesized DNA that cannot be subdivided into smaller component parts. BBa\_R0051 is an example of a promoter basic part. Basic parts have the role of sources in the Lexis setting. A ``composite part'' is a functional unit of DNA consisting of two or more basic parts assembled together. BBa\_I13507 is an example of a composite part, consisting of four basic parts ``BBa\_B0034 BBa\_E1010 BBa\_B0010 BBa\_B0012''. 
The dataset we analyze is the set of all composite parts submitted to the registry from 2003 to 2017. In this dataset, the composite parts are represented by the string of their basic parts (i.e., a non-dividing representation). The sequence of iGEM composite parts can be considered as a sequence of target strings over a set of sources (i.e., basic parts). 
We have acquired the iGEM data from \url{https://github.com/biohubx/igem-data}. All the BioBrick parts were crawled until Dec 28th 2017.
In Table \ref{stats}, the preliminary statistics about the dataset are listed.
\begin{table}[h]
\centering
\caption{Basic statistics on iGEM dataset during 15 years (2003-2017)\label{stats}}
\begin{tabular}{|c|c|c|c|}
\hline
{\bf \# Sources} & {\bf \# Targets} & {\bf Total Length} & {\bf Min/Max Target Length}
\\\hline
7,889 & 18,394 & 107,022 & 2 / 100
\\\hline
\end{tabular}
\end{table}
The dataset mostly presents targets of small length. The top 5 categories having the highest fraction of the targets belongs to those of length 5, 2, 3, 4 and 6, accounting for more than 70\% of the dataset. Less than 10\% of the targets have a length of more than 10.


\subsection{Considering Annual Batches of Targets}

The iGEM competition is conducted annually. Hence, it is reasonable to consider the sequences of targets as annual batches of targets arriving each year.
This consideration is in line with the incremental design process in Evo-Lexis.

\begin{figure}[h!]
\centering
\subfloat[
\label{y-st}]{
\includegraphics[trim= 1cm 6cm 2cm 6cm, clip, width=.35\textwidth]{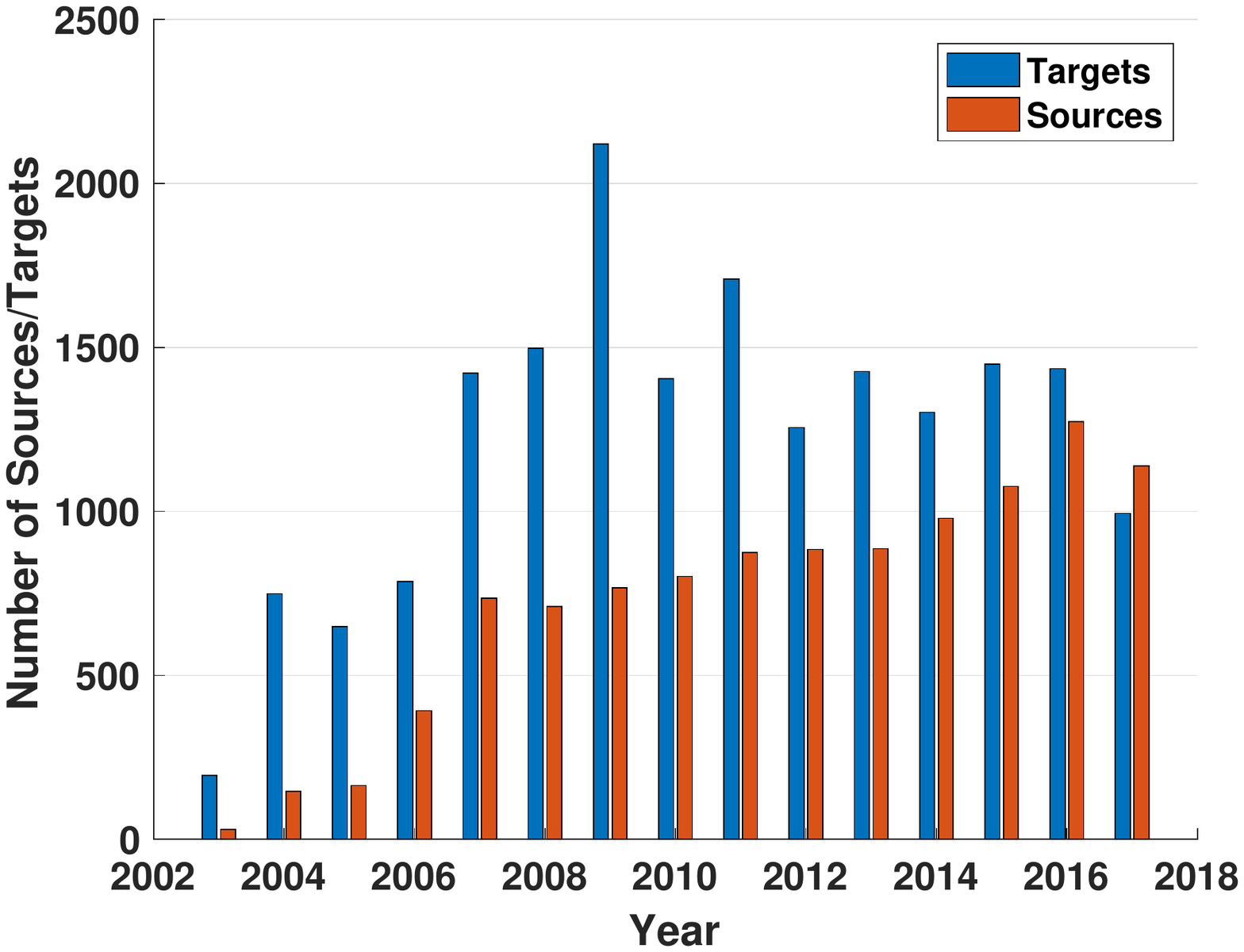}
}
\subfloat[
]{
\includegraphics[trim= 1cm 6cm 2cm 6cm, clip, width=.35\textwidth]{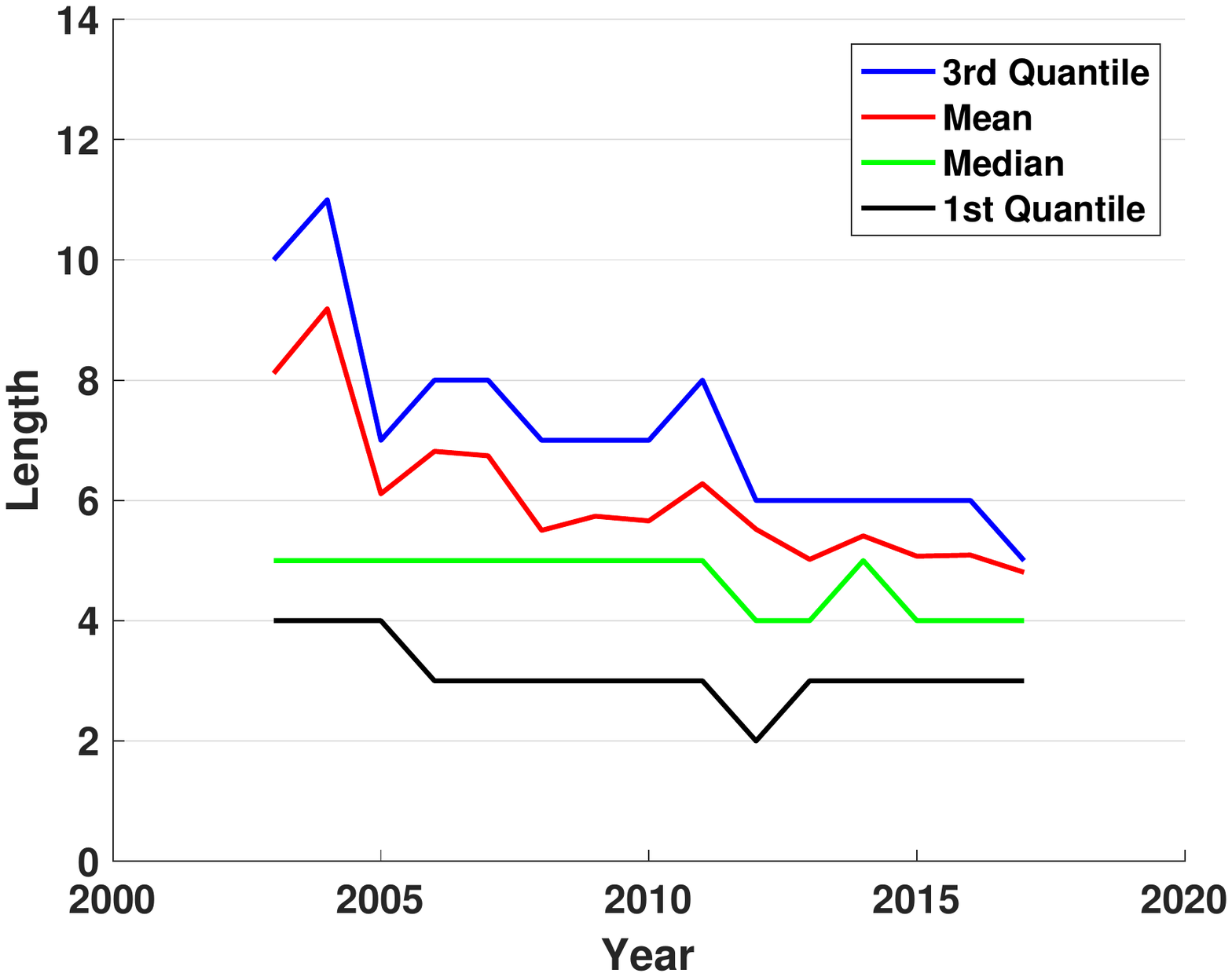}
}
\\
\subfloat[Source set similarity in years \label{sy-y}]{
\includegraphics[trim= 1cm 6cm 2cm 6cm, clip, width=.35\textwidth]{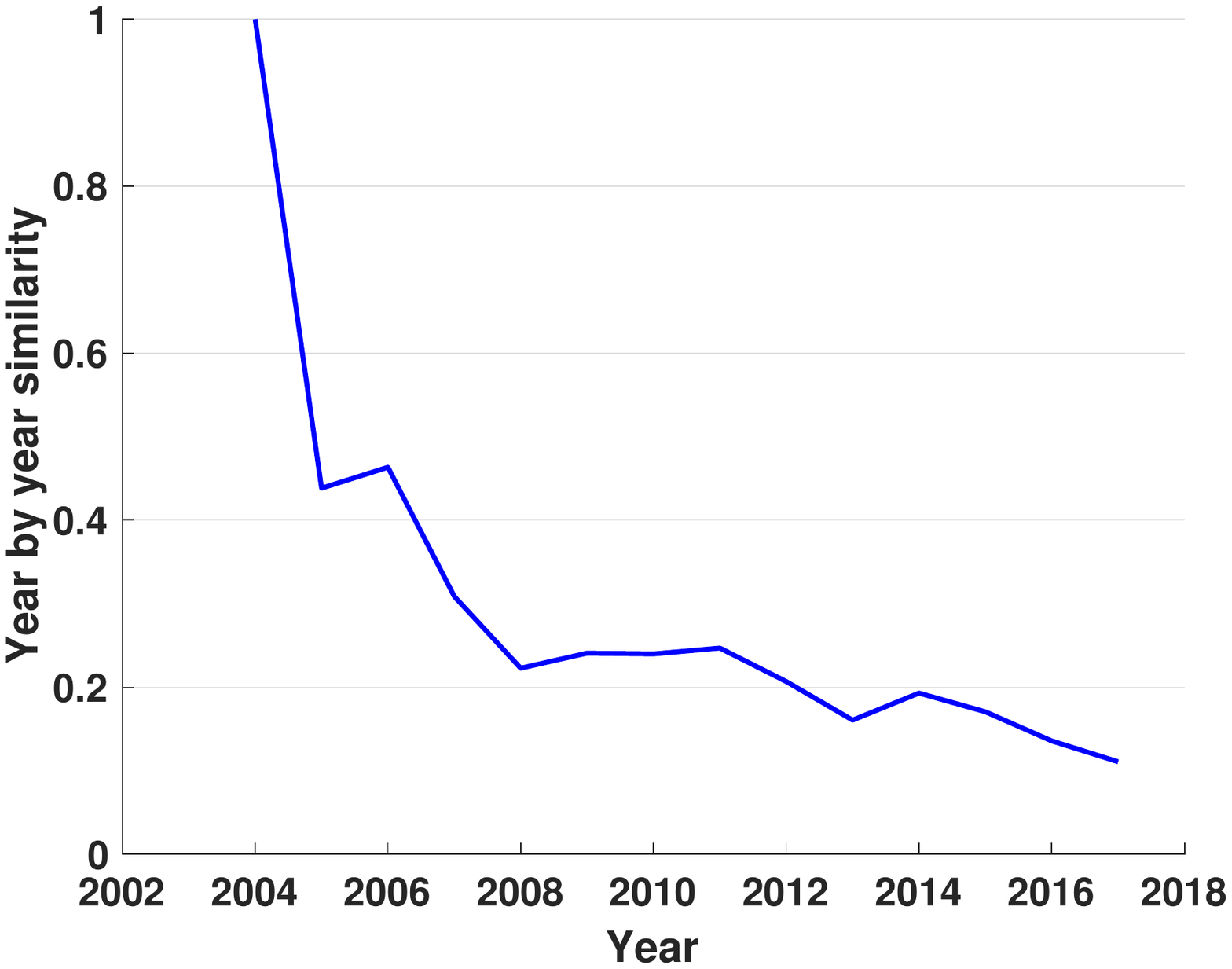}
}
\subfloat[\label{src-reuse} PDF of reuse of the sources per year.]{
\includegraphics[trim= 0.1cm .5cm .1cm .5cm, clip, width=0.65\textwidth]{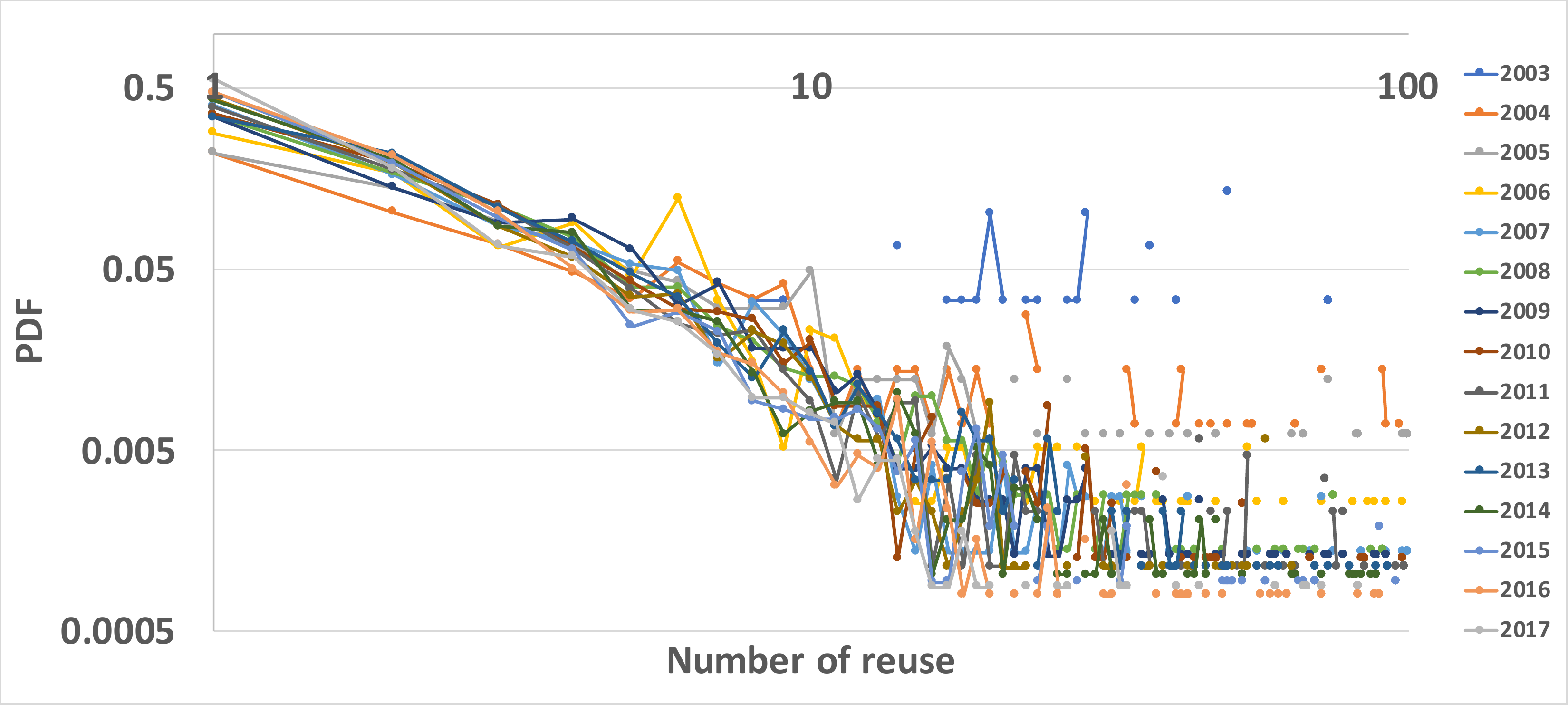}
}
\caption{Statistics of iGEM dataset when considered as yearly batches. Number of reuse is the number of times a source appear in a target in each year.\label{y-stats}}
\end{figure}

To show some differences between iGEM and Evo-Lexis, in Fig. \ref{y-stats}, we can see how the number of sources, the number of targets, length statistics and source reuse statistics change over time.
We can make the following observations from these figures:
\begin{enumerate}
\item
The number of sources increases, where it was constant in Evo-Lexis.
\item
In the first four years, the number of targets per year is noticeably small. Later on, the number of targets increases up to 2,000 and then fluctuates around 1,000 to 1,300 targets per year. In Evo-Lexis, the number of targets per batch is constant and they all have the same length.
\item
The mean and median of target lengths stay in the same range ($\in [5,7]$) during all 15 years.
\item
The reuse of sources (except for the beginning years) is extremely skewed in all years: few sources are used much more often than most of the sources (Fig. \ref{src-reuse}). In Evo-Lexis, all sources are equally likely.
\end{enumerate}
In the following sections, we show that how these differences between iGEM dataset and Evo-Lexis cause differences between the resulting Lexis-DAGs.

\section{Analysis of iGEM Dataset in Evo-Lexis Framework}
\label{sec:analysis}
From this section on, we compare the results over iGEM with the results gathered from Evo-Lexis in \cite{siyari2018}. We refer the reader to \cite{siyari2018} for details of the model and parameter settings.
\subsection{Lexis-DAG Cost Analysis}
\begin{figure}[h!]
\centering
\subfloat[The cost evolution in iGEM]{
\includegraphics[trim= 1cm 6cm 2cm 6cm, clip, width=0.45\textwidth]{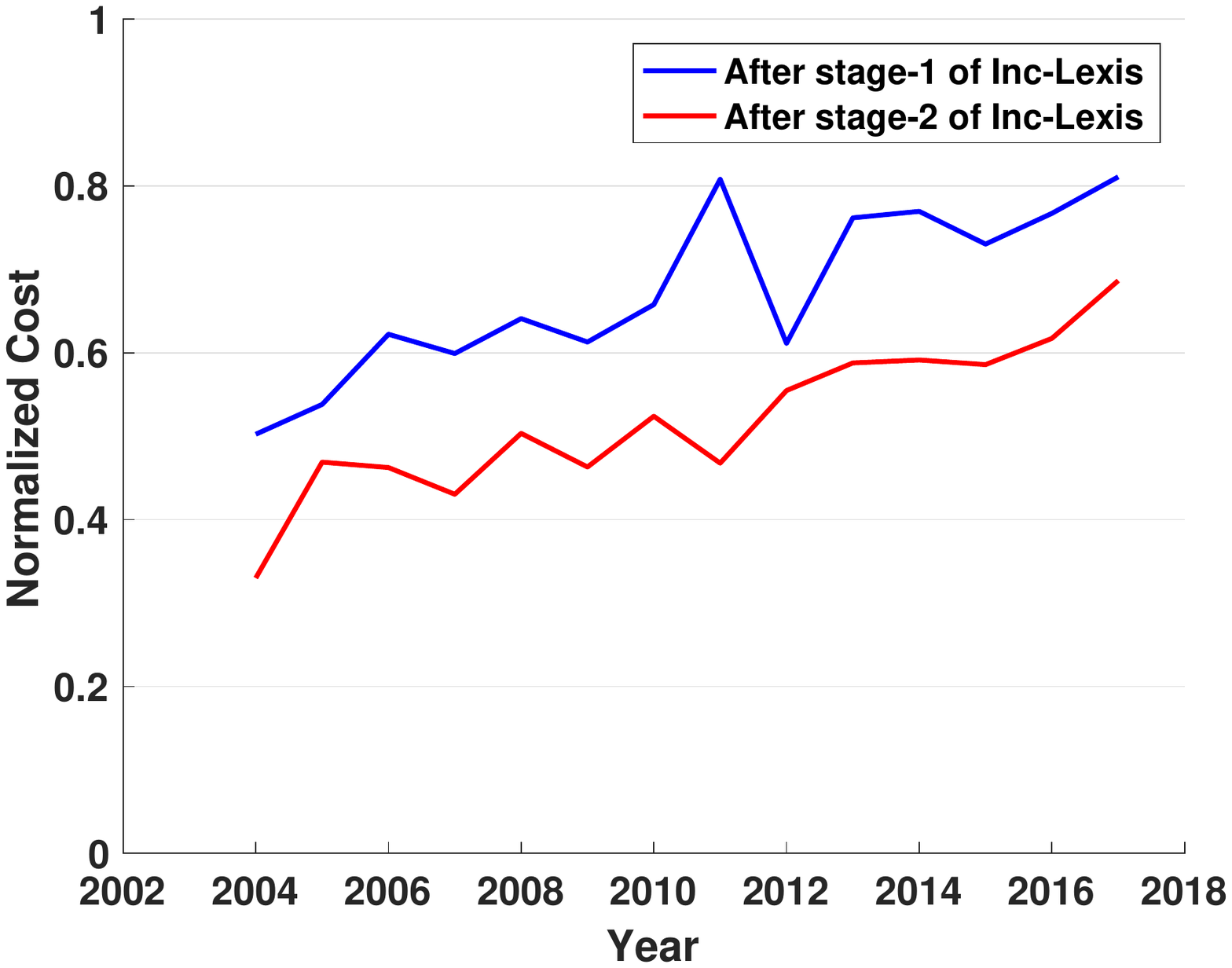}
}
\subfloat[The cost evolution in Evo-Lexis]{
\includegraphics[trim= 1cm 6cm 2cm 6cm, clip, width=0.45\textwidth]{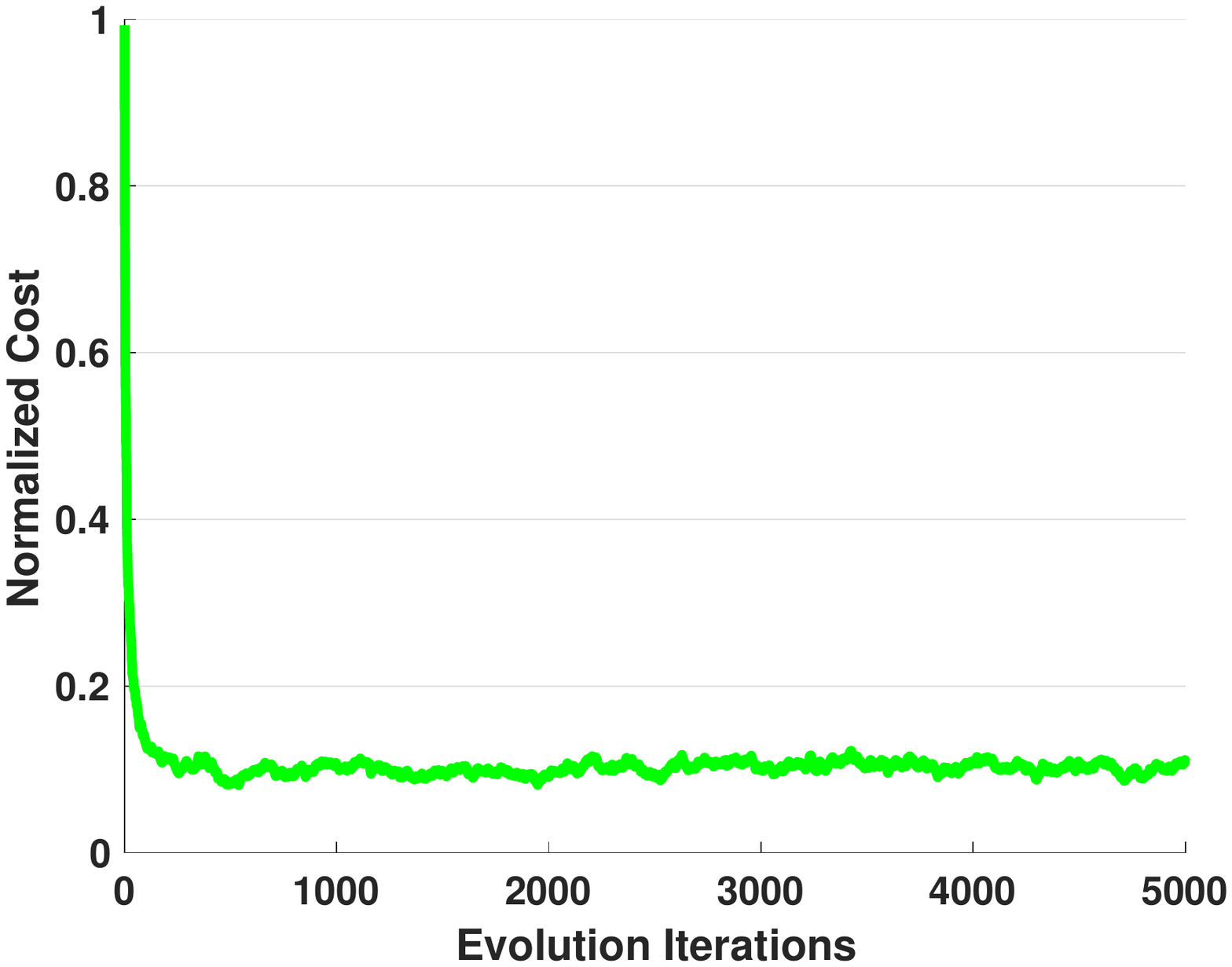}
}
\caption{\label{inc-cost} Comparison of cost evolution in iGEM and Evo-Lexis (from \cite{siyari2018})}
\end{figure}

In this section, we observe how cost efficient the Lexis-DAGs over the iGEM dataset are. We consider an incremental setting similar to Evo-Lexis: In the first year, a clean-slate Lexis-DAG is constructed over the targets of that year. For the targets of the subsequent years, an incremental Lexis-DAG is constructed. Fig. \ref{inc-cost} shows how the normalized cost of the Lexis-DAGs varies over the years on iGEM. We observe major differences with Evo-Lexis; in Evo-Lexis the normalized cost remains almost constant.

To investigate the reasons for the above observations, in the same Fig. \ref{inc-cost}, we also track the cost reduction performance of the two stages of \textsc{Inc-Lexis} for each batch (as a reminder, in stage-1, we reuse intermediate nodes from previous Lexis-DAG and in stage-2, we further optimize the hierarchy using \textsc{G-Lexis}). This experiment is done due to our interest in seeing how much stage-1 of \textsc{Inc-Lexis} contributes to the cost reduction on iGEM. There are two observations that we can make:
\begin{enumerate}
\item 
In most batches, more than 50\% of the cost reduction is achieved by the stage-1, i.e., reuse stage. The contribution of stage-2 of \textsc{Inc-Lexis} is roughly constant throughout years. This suggests that iGEM targets reuse a significant amount of sequences from previous years in their own submissions.
\item
There is an increasing trend in the normalized cost after stage-1. This observation means that the contribution of the reuse stage in \textsc{Inc-Lexis} decreases over the years. As mentioned, the contribution of stage-2 stays mostly constant. Hence, we can relate the increasing trend of the normalized cost to the fact that the amount of reuse reduces from year to year.
\end{enumerate}

\label{new-src}
We can find the root-cause of the decrease of reuse over time on iGEM to the increase of the size of the set of sources. We have observed in Fig. \ref{y-st} that there are many new sources that get introduced over the years. One of the requirements for reuse from one batch to another in Evo-Lexis is the fact that the set of sources does not drastically change (in fact it is constant in the Evo-Lexis framework). To investigate whether this is true in iGEM, we check the ratio of the sources from one year to the next that remain the same. Specifically, if we have $y_2 = y_1 + 1$, and if $S_{y_1}$ \& $S_{y_2}$ are the set of sources in year $y_1$ \& $y_2$ respectively, we check the ratio: $\frac{|S_{y_1} \cap S_{y_2}|}{|S_{y_1}|}$. This ratio, i.e., \emph{year-by-year similarity}, is the fraction of sources that remain from the previous year. Fig. \ref{sy-y} shows how this ratio changes from year to year. By year 2008, the ratio drops significantly to a value around 0.2 which means around 80\% of the sources from the previous year are not reused. This reduces the amount of reuse that is possible in the iGEM dataset. The introduction of new sources is also propagated in individual targets. 
As time progresses, there is a higher probability to use more than $X$ number of new sources per target. This observation is a further obstacle for reuse, especially given that the targets in iGEM are often short (5-7 subparts). Following the increase of the normalized cost, Fig. \ref{d-l} shows that the DAGs get less deep and have lower average node length as time progresses. Overall, the results of this section show a number of differences between iGEM and  Evo-Lexis:
\begin{figure}[h!]
\centering
\subfloat[Average depth]{
\includegraphics[trim= 1cm 6cm 2cm 6cm, clip, width=.4\textwidth]{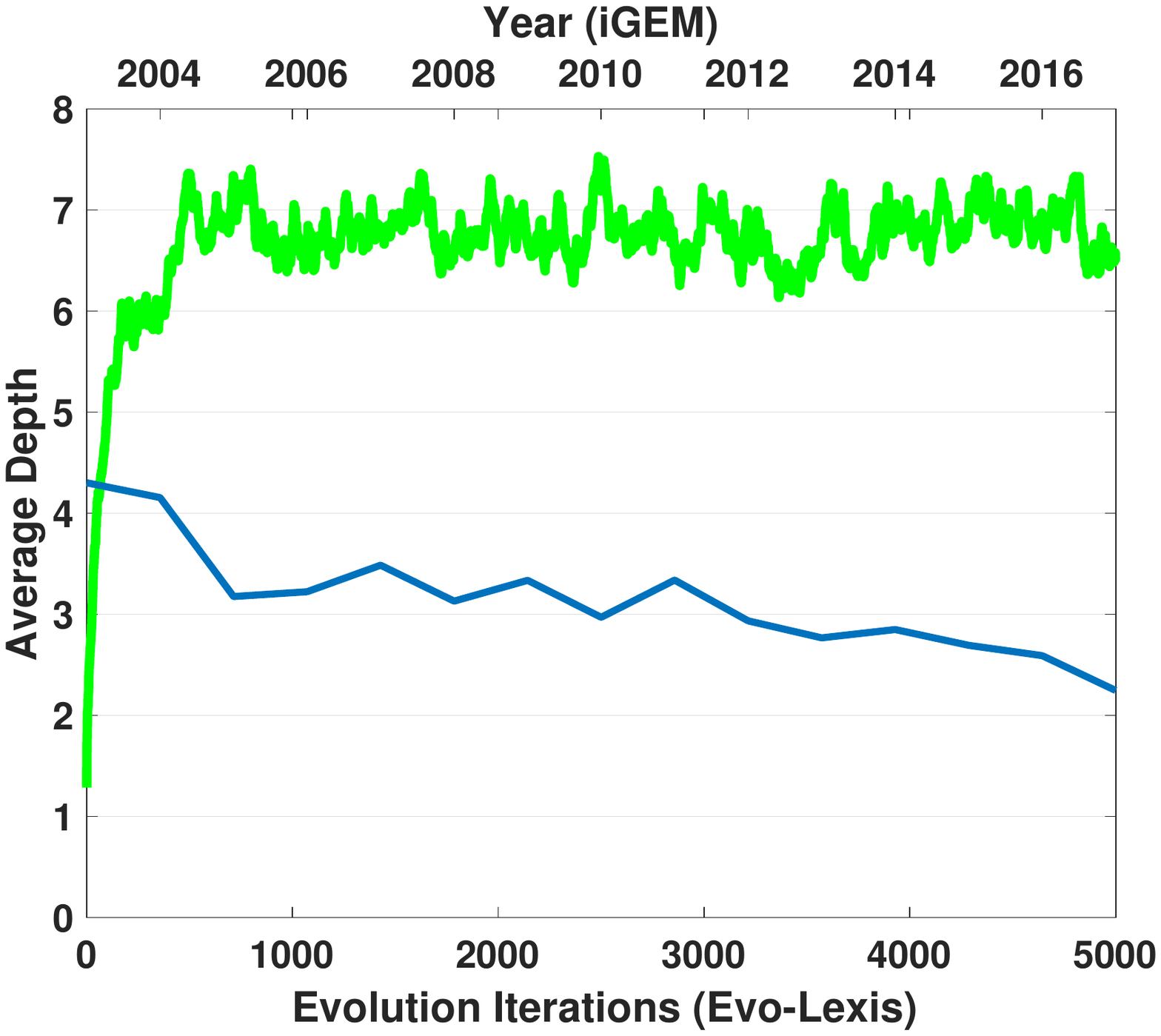}
}
\subfloat[Average node length]{
\includegraphics[trim= 1cm 6cm 2cm 6cm, clip, width=.4\textwidth]
{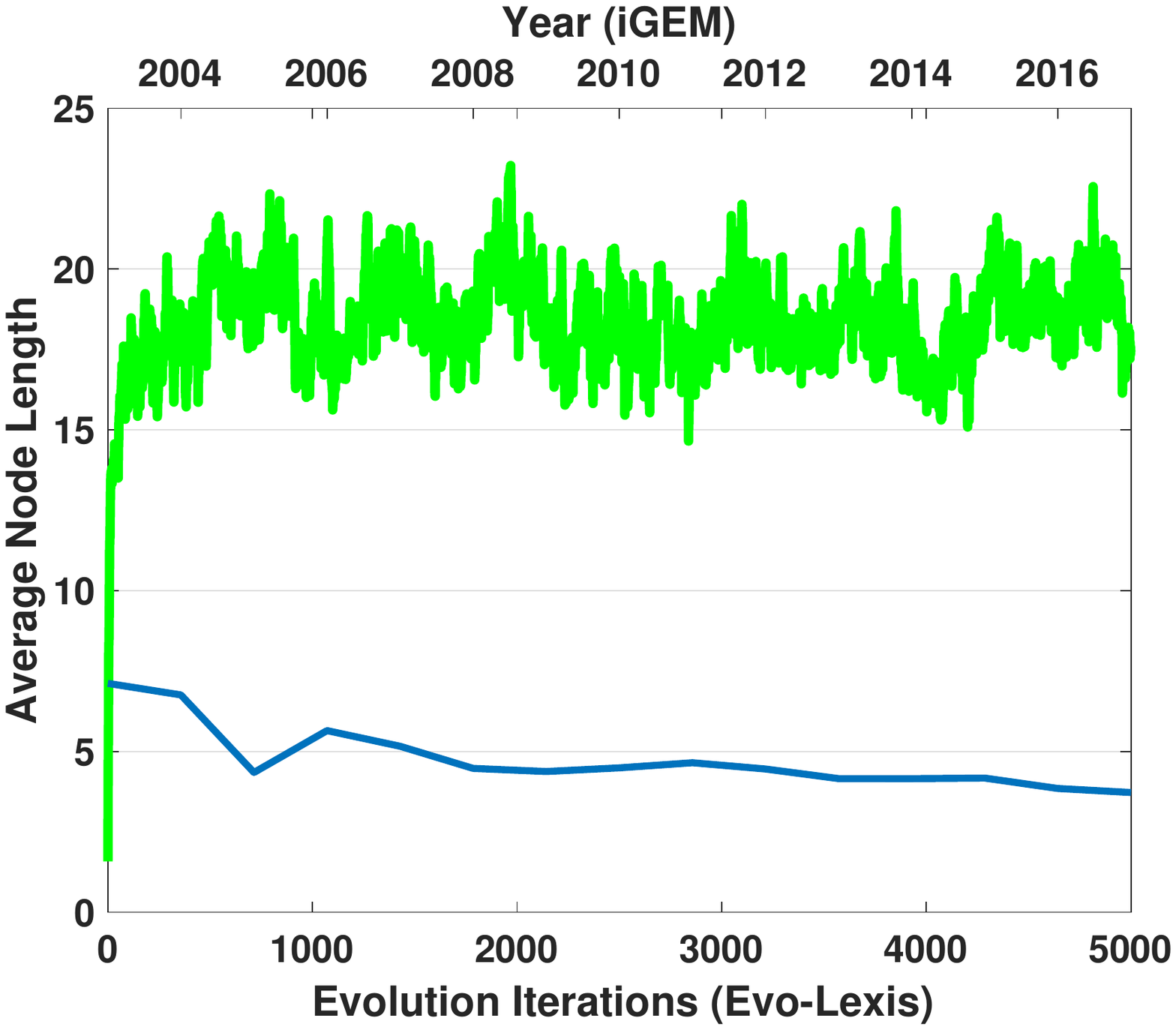}
}
\caption{Average depth and node length in iGEM and Evo-Lexis (in green, \cite{siyari2018})
\label{d-l}
}
\end{figure}
\begin{enumerate}
\item In iGEM, the set of sources in each year has low similarity to the previous years, while in Evo-Lexis the source set is constant. The high amount of churn in the set of sources is the primary reason for the lower reuse in iGEM data compared to Evo-Lexis. The fact that the targets are shorter is another factor for iGEM's lower potential for reuse of longer intermediate nodes.
\item
The normalized cost, depth and average node length are all lower in iGEM due to the reduced reuse potential as discussed above.
\end{enumerate}
\subsection{Hourglass Effect in iGEM}
\begin{figure}[h!]
\centering
\subfloat[Core Size \label{coreSize}]{
\includegraphics[trim= 1.3cm 6cm 2cm 6cm, clip, width=.3\textwidth]{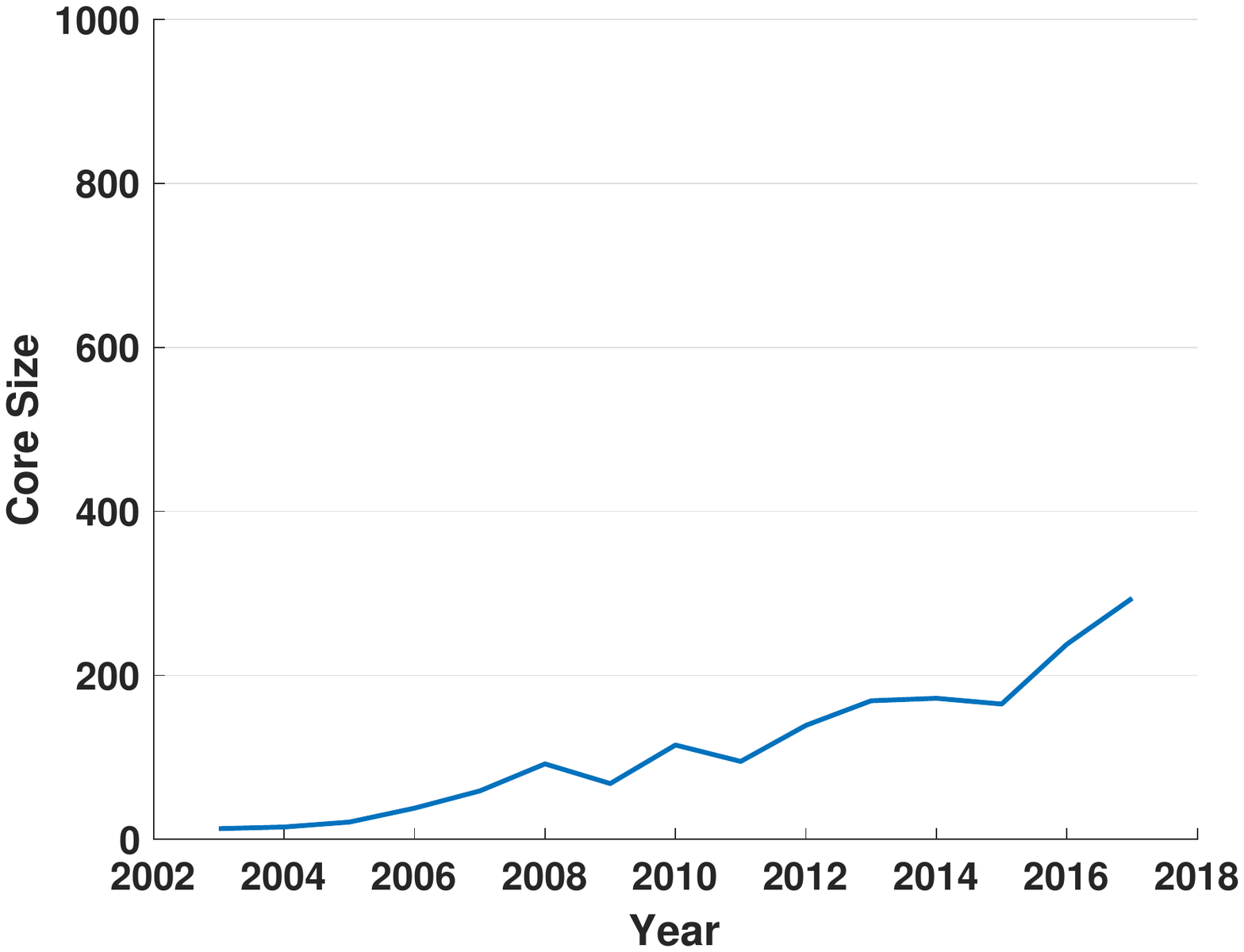}
}
\subfloat[Flat DAG Core Size\label{flat}]{
\includegraphics[trim= 1.3cm 6cm 2cm 6cm, clip, width=.3\textwidth]
{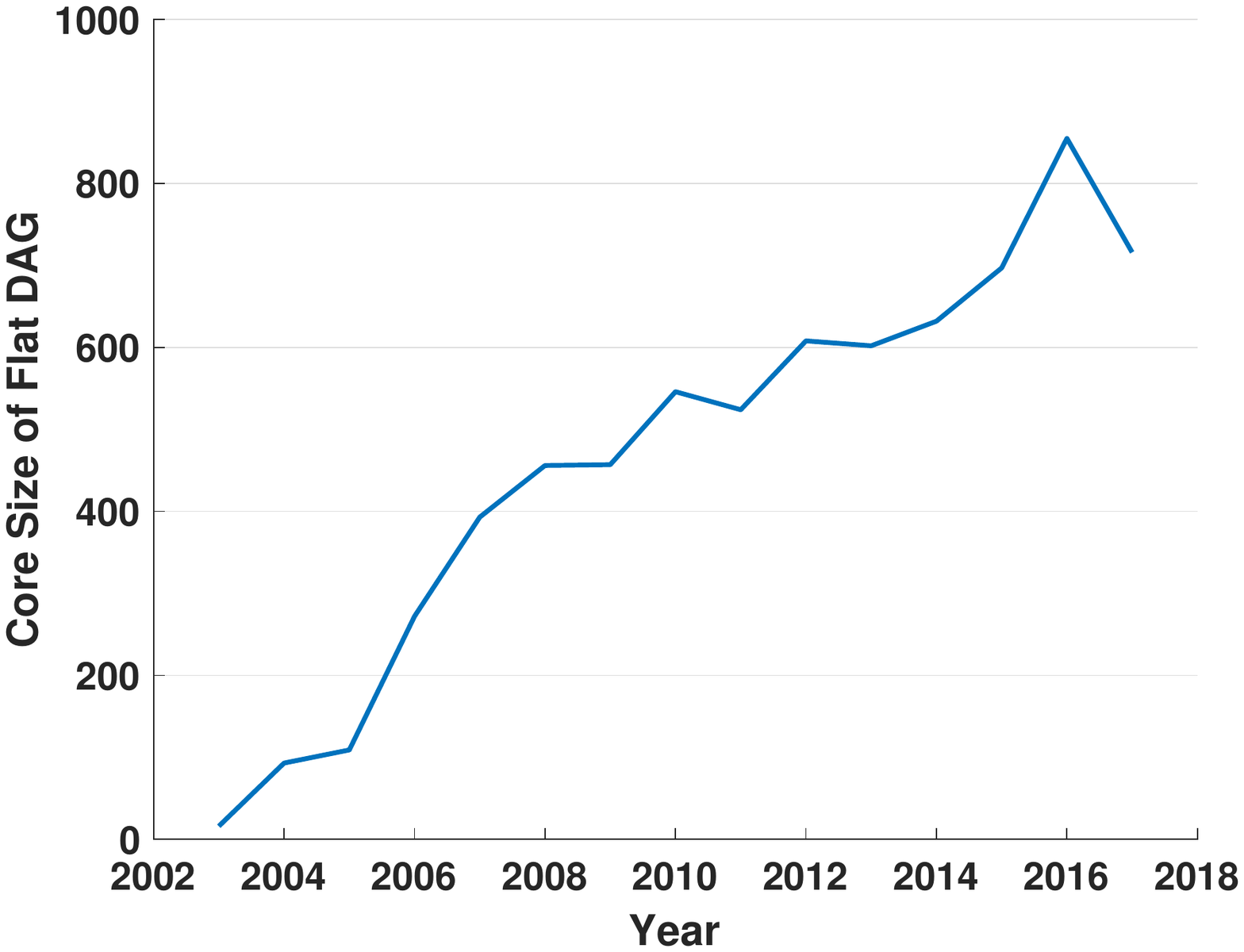}
}
\subfloat[H-score\label{hscore}]{
\includegraphics[trim= 1.3cm 6cm 2cm 6cm, clip, width=.3\textwidth]
{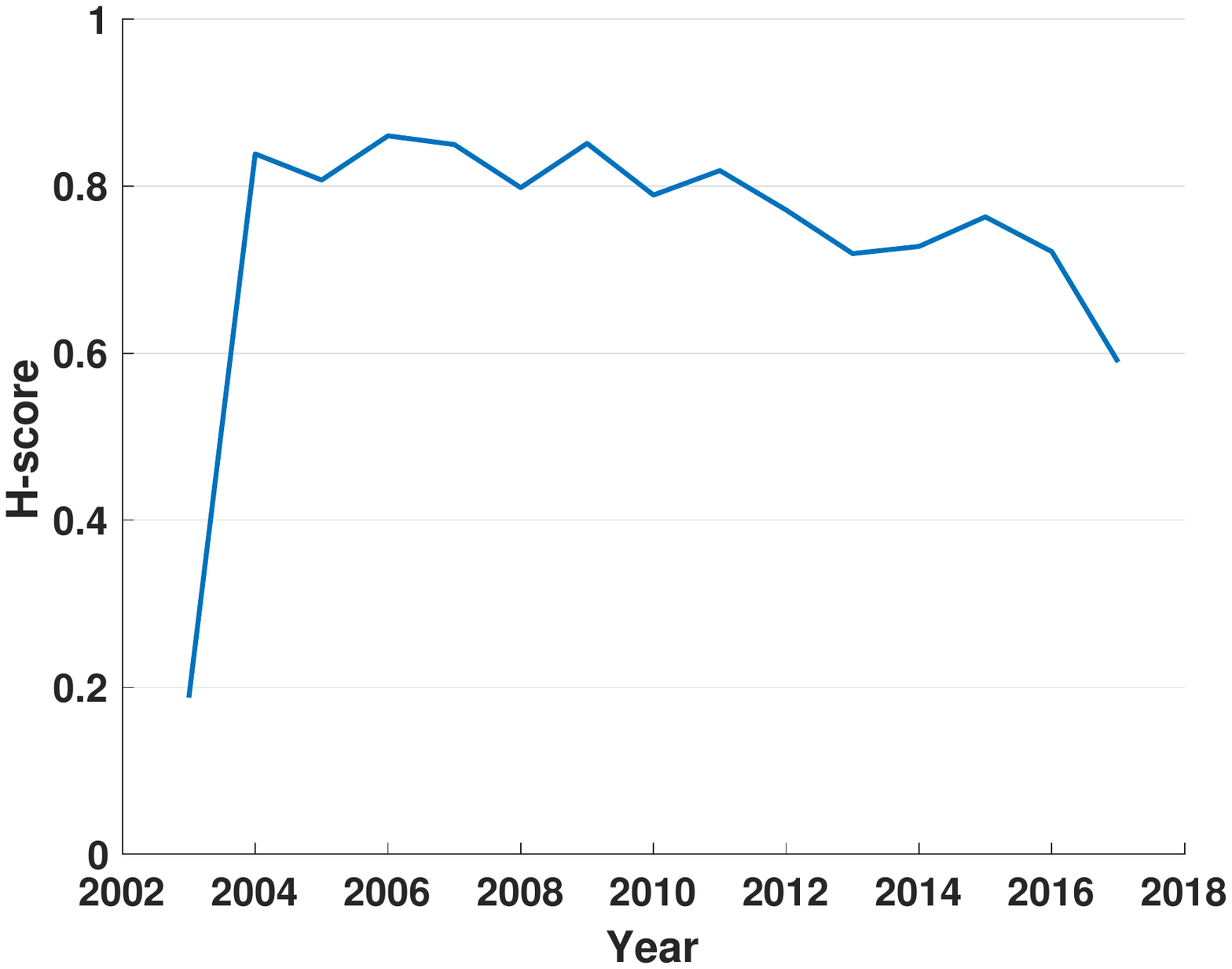}
}
\\
\subfloat[Core Size \label{coreSizee}]{
\includegraphics[trim= 1.3cm 6cm 2cm 6cm, clip, width=.3\textwidth]{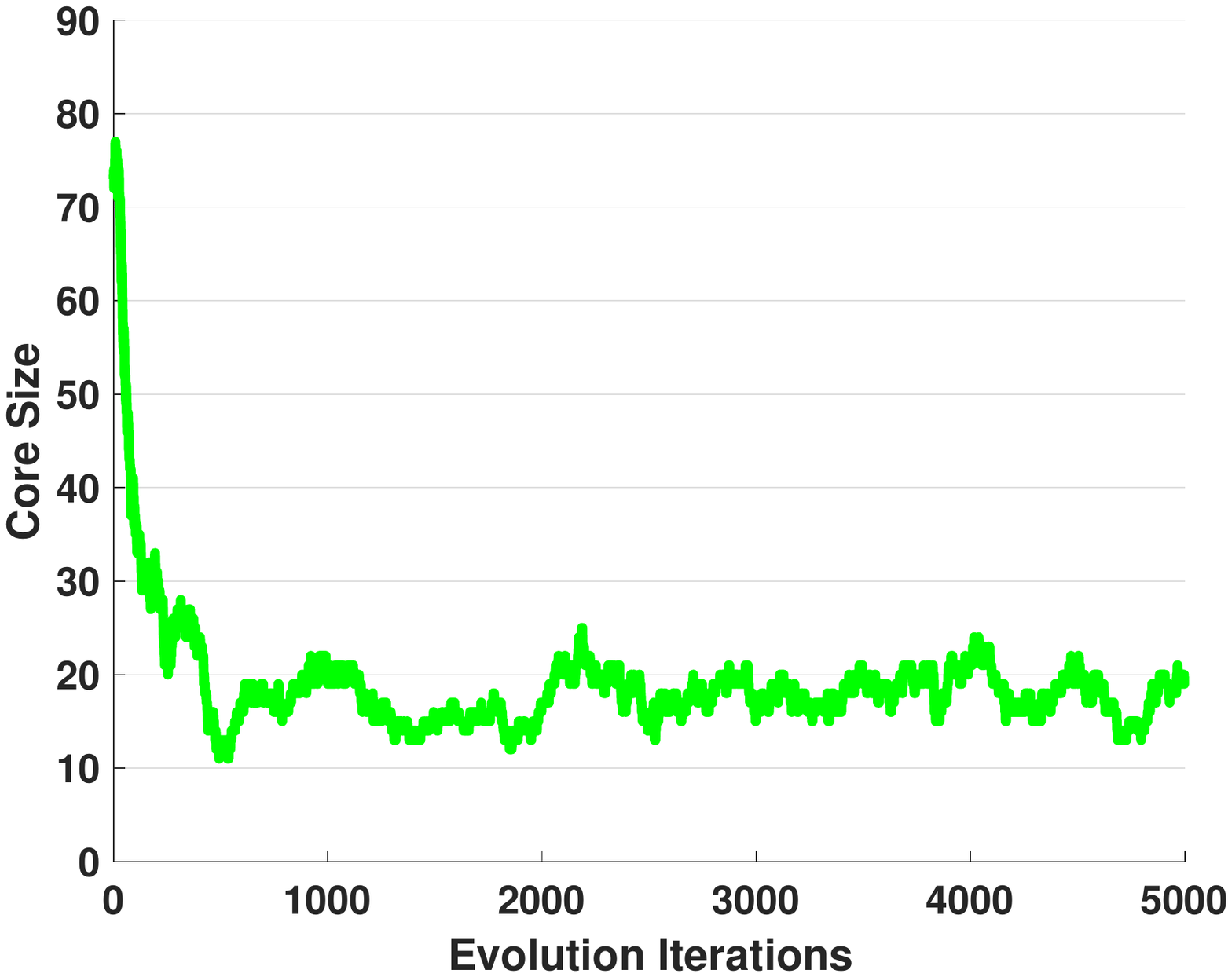}
}
\subfloat[Flat DAG Core Size\label{flate}]{
\includegraphics[trim= 1.3cm 6cm 2cm 6cm, clip, width=.3\textwidth]
{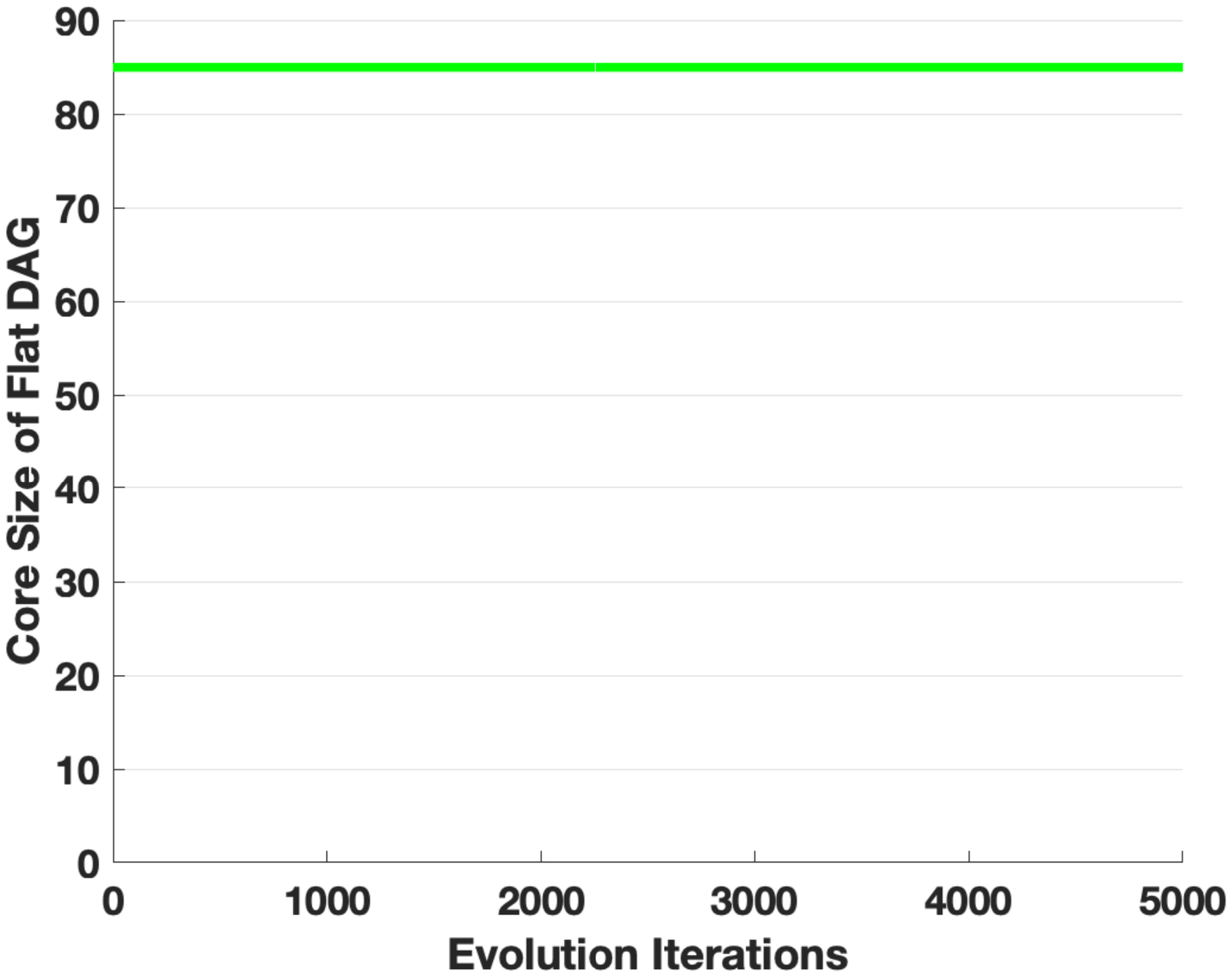}
}
\subfloat[H-score\label{hscoree}]{
\includegraphics[trim= 1.3cm 6cm 2cm 6cm, clip, width=.3\textwidth]
{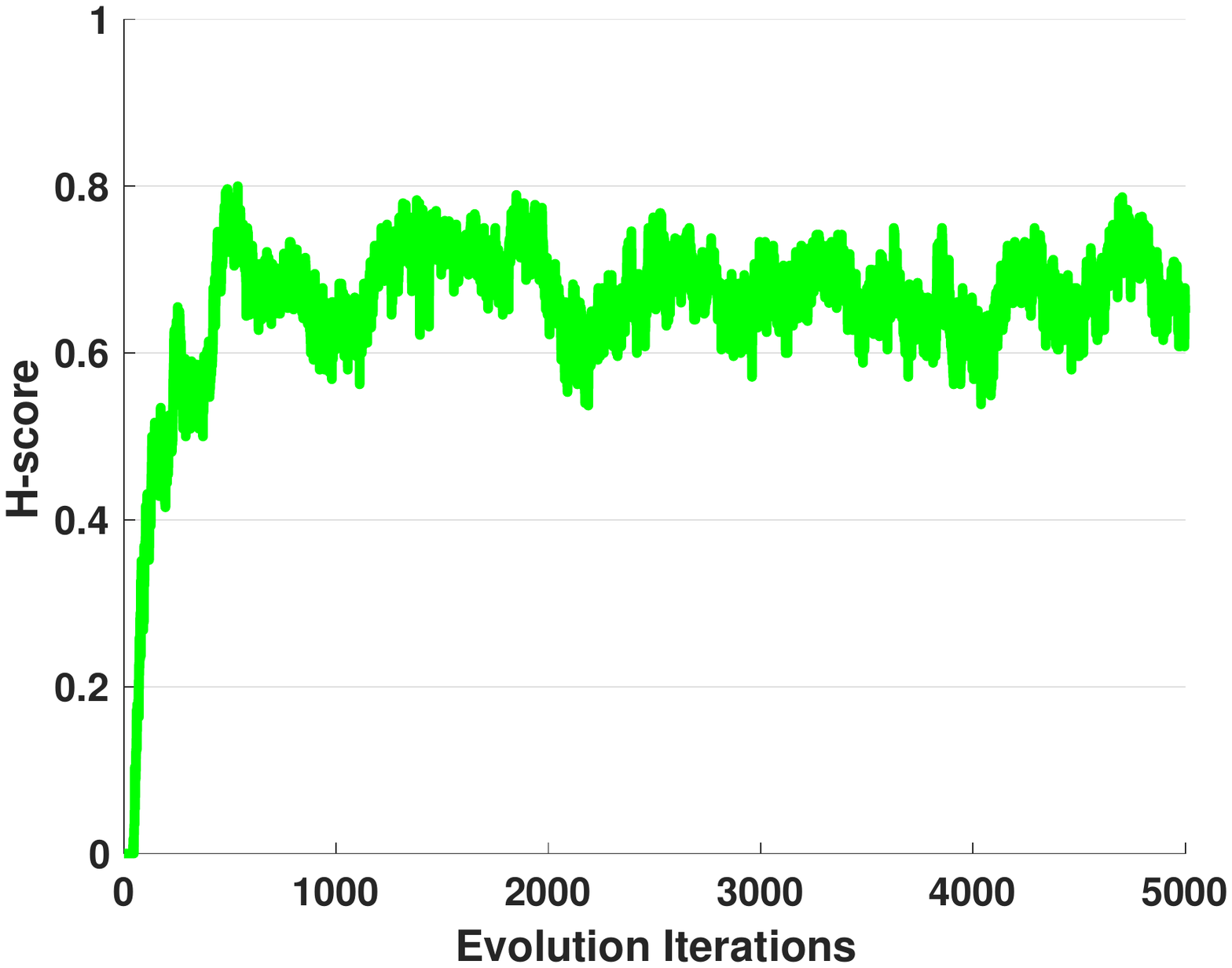}
}
\caption{Cores in iGEM and Evo-Lexis (bottom, \cite{siyari2018})  ($\tau=0.85$).\label{core-hscore}
}
\end{figure}

The following results in this section show 
that in all years, there is a small number of core nodes in the iGEM Lexis-DAGs.
Fig. \ref{core-hscore} shows that such small cores make the topology of iGEM Lexis-DAGs consistent with an hourglass organization (high H-score values - more than 0.6 in Fig. \ref{hscore}). In Evo-Lexis, we observe similar values of H-score for DAGs constructed using synthetic data. As observed, although the core size increases in iGEM  over time, we see a steeper increase in the size of the flat DAG's core mostly due to the increase in set of sources. In Evo-Lexis, the core size shows a decreasing trend while the size of the core of the flat DAG does not significantly change, reflecting similarly high H-score values as in iGEM. Overall, we can see that the topology of the Lexis-DAGs in iGEM data is in line with the Evo-Lexis model, although the bias in selection of cost-saving nodes is not sufficiently large to cause a non-increasing normalized cost.

\subsection{Diversity among iGEM Targets}
Another question is the degree of diversity among the targets of iGEM over time. 
%
We define the concept of \emph{Normalized Diversity} as follows: Suppose we have a set of strings $T=\{t_1, t_2, ..., t_n\}$. The goal is to provide a single number that quantifies how dissimilar these elements are to each other.
\begin{itemize}
\item 
We first identify the \emph{medoid} $\mathcal{M}_T$ of the set $T$, i.e., the element that has the lowest average distance from all other elements. We use Levenshtein distance as a measure of distance between targets: $\mathcal{M}_T=arg~min_{m \in T} \sum_{t \in T}LD(t,m)$.
\item
To compute how diverse the elements are with respect to each other, we average  the normalized distance of all elements from the medoid (distance is normalized by the maximum length of the two sequences in question). We call this measure $\sigma_T$, the \emph{Normalized Diversity} of set $T$. The bigger the metric, the more diverse a set of strings is: 
$\sigma_T=\frac{\sum_{t \in T}\frac{LD\left[t,\mathcal{M}_T\right]}{max(|t|,|\mathcal{M}_T|)}}{|T|}$.
\end{itemize}

\begin{figure}[h!]
\centering
\subfloat[Target diversity in iGEM]{
\includegraphics[trim= 1cm 6cm 2cm 6cm, clip, width=.33\textwidth]{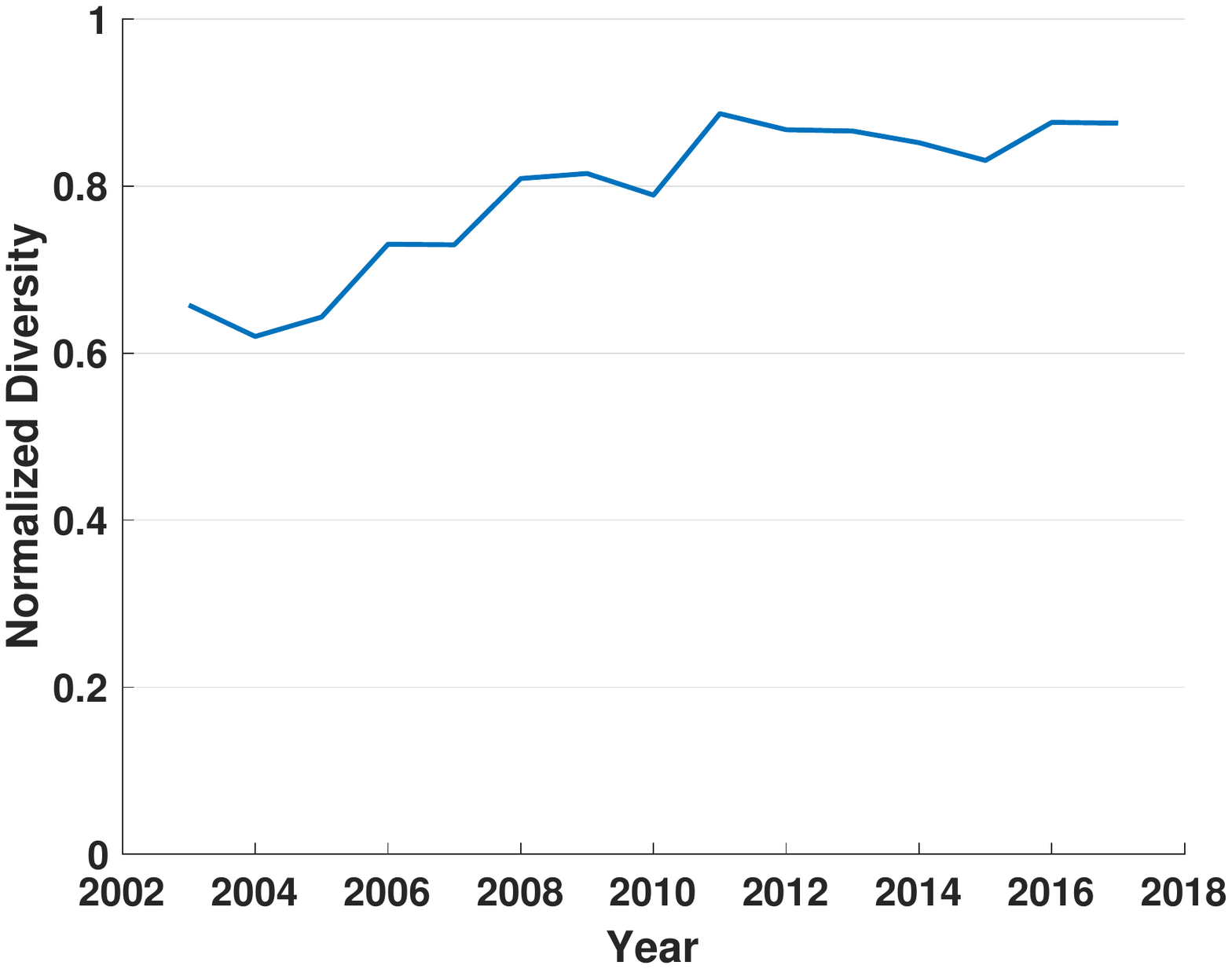}
\label{diversity}
}
\subfloat[Core stability in iGEM]{
\includegraphics[trim= 1cm 6cm 2cm 6cm, clip, width=.33\textwidth]{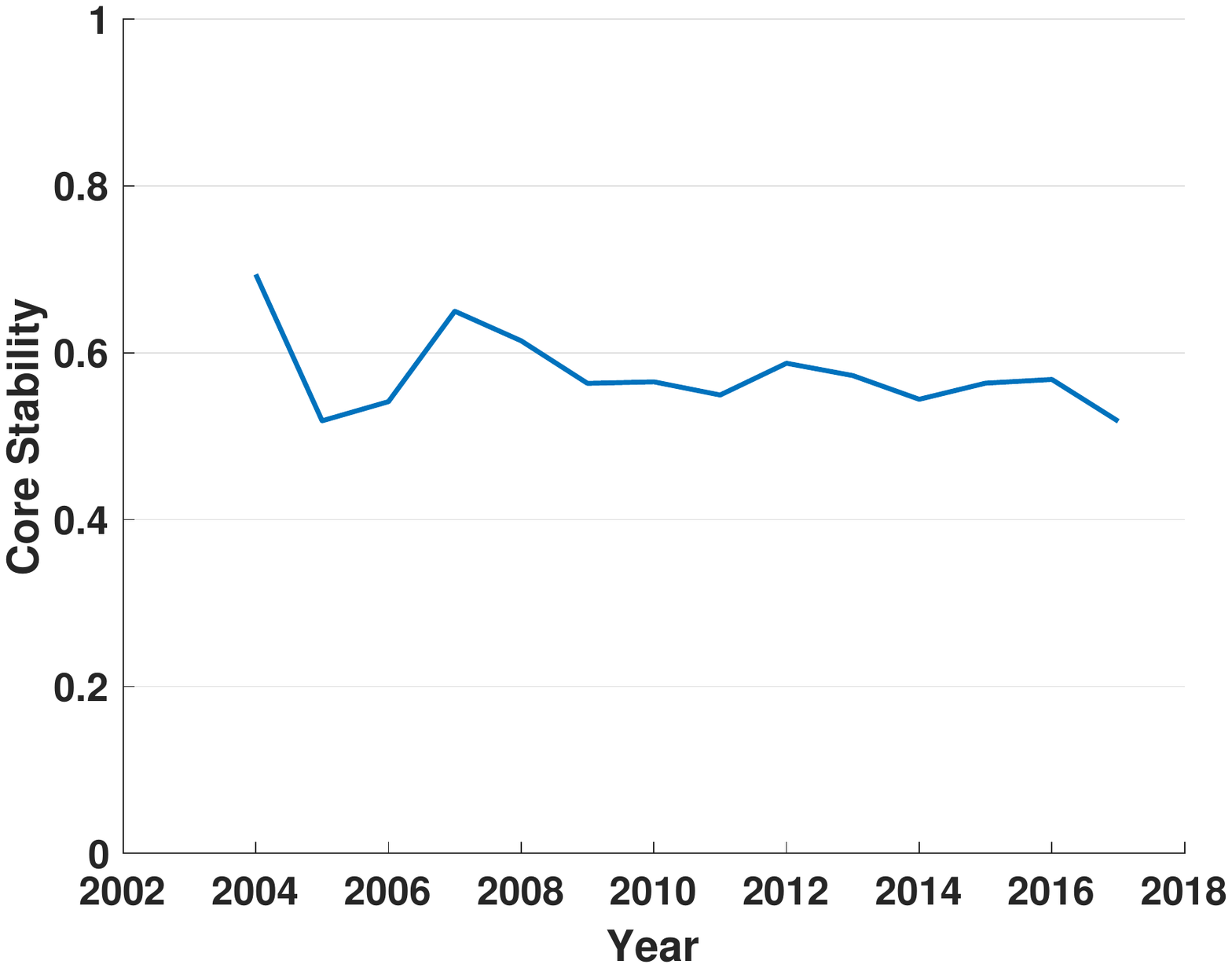}
\label{stability}
}
\subfloat[Core stability in Evo-Lexis]{
\includegraphics[trim= 1cm 6cm 2cm 6cm, clip, width=.33\textwidth]{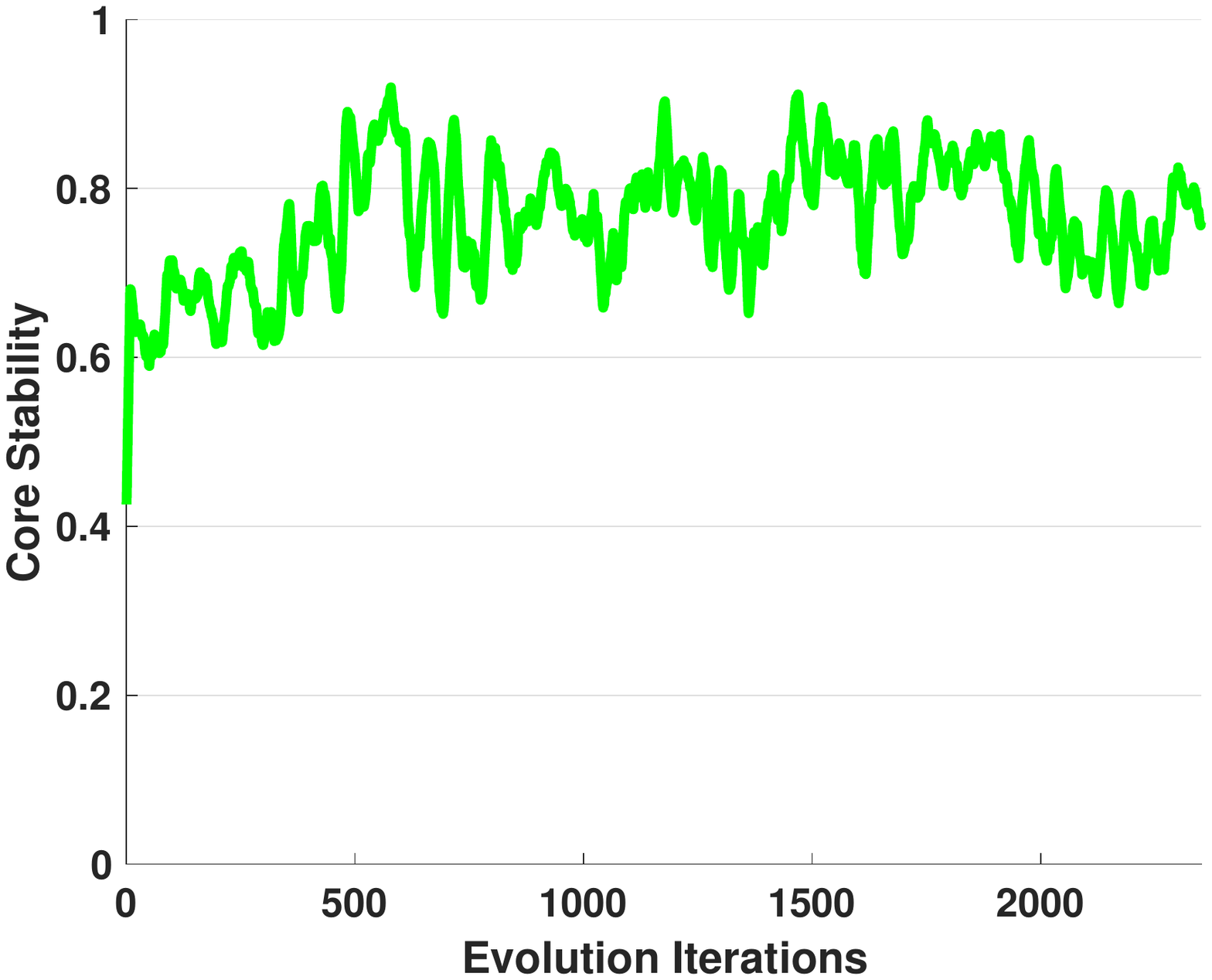}
\label{stability}
}
\caption{
Target diversity and core stability in iGEM over time.
\label{diversity-stab}}
\end{figure}

Fig. \ref{diversity-stab} shows that the normalized diversity metric has a value of more than 0.5 throughout time and reaches up to 0.8 (this means that on average 50\% to 80\% of a target should be changed so that a target is converted to another in the set of targets in each year). Although such values of diversity are in line with Evo-Lexis, it is understandable that the diversity in iGEM is also partially impacted (towards higher values) by the introduction of new sources discussed before. Because of this reason, and the fact that the diversity is measured in a slightly different way in \cite{siyari2018}, we do not show a direct comparison in Fig. \ref{diversity-stab}.


\subsection{Core Stability in iGEM Lexis-DAGs}
We have already defined the core size and the H-score. Here we define an additional metric, related to the stability of the core across time.

We track the stability of the core set
by comparing two core sets at two different times. A direct comparison of the core sets via the Jaccard index leads to poor results. The reason is that often 
the strings of the two sets 
are similar to each other but not completely identical.

Thus, we define a generalized version of Jaccard similarity that we call \emph{Levenshtein-Jaccard Similarity}:
\begin{itemize}
\item
Suppose we aim to compute the similarity of two sets A and B of strings. We define the mapping $A\rightarrow B$ where every element $a \in A$ is mapped to the most similar element $b \in B$. We also define the mapping $B \rightarrow A$ from every element $b \in B$ to the most similar element $a \in A$:
\begin{equation}
\begin{cases}
A\rightarrow B = \{(a,b)~s.t.~a \in A~\&~b \in B~\&~b = arg~max_{x \in B} Sim(a,x)\}
\\
B\rightarrow A = \{(b,a)~s.t.~a \in A~\&~b \in B~\&~a = arg~max_{x \in A} Sim(b,x)\}
\end{cases}
\end{equation}
where $Sim(a,b)$ is the similarity of $a$ to $b$ and is calculated as: $Sim(a,b) = 1-\frac{LD(a,b)}{max(|a|,|b|)}$.
Notice that $max(|a|,|b|)$ is the maximum value of Levenshtein distance between $a$ and $b$. This consideration ensures that if $a=b$ then $Sim(a,b) = 1$, and if $a$ and $b$ have the maximum distance then $Sim(a,b)=0$.
\item
Considering both $A \rightarrow B$ and $B \rightarrow A$, we get the union of the two mappings and define the Levenshtein-Jaccard similarity as follows:
\begin{equation}
LevJac(A,B) = \frac{\sum_{(a,b) \in A \rightarrow B} Sim(a,b) + \sum_{(b,a) \in B \rightarrow A} Sim(b,a)}{(|A|+|B|)}
\label{stability-eq}
\end{equation}
We can see that if $A=B$ (all weights are equal to one) then $LevJac(A,B) = 1$.
Also if none of the elements in $A$ are similar to $B$ (all the element pairs take zero similarity value), then $LevJac(A,B) = 0$.
\end{itemize}


As the results in Fig. \ref{stability} show, the core set in iGEM DAGs have relatively high values of the core stability measure (Eq. \eqref{stability-eq}), close to the values we observed in Evo-Lexis. This means that the core nodes stay similar across time, and there are no sudden changes in the content of the core set. 
One reason for this stability is that the set of core nodes includes several sources, and many of core sources get transferred to the next year. 


Additionally, every year the focus of the iGEM designers is on specific parts, most of which are of high path centrality. For example, ``BBa\_B0010 BBa\_B0012'' (the most widely used ``terminator'' part) and ``BBa\_B0034'' are almost always the top-2 central nodes (with the exception of year 2011). Also, some sources such as ``BBa\_R0011'', always appear in the top-20 nodes in the core set. Remember that Fig. \ref{src-reuse} shows that the reuse distribution of sources is highly skewed.
In summary, the stability of the core set in iGEM is caused by the same reason with Evo-Lexis, which is the bias and selectivity towards using a specific set of nodes  in consecutive years.

\section{Conclusions}
\label{sec:conclusion}
iGEM is a dataset that satisfies the basic assumption of Evo-Lexis framework: a sequence of target strings with potential temporal reuse of previously introduced substrings. Because of this compatibility, we chose to use  this dataset in a case-study and contrast its qualitative properties with Evo-Lexis.
We can summarize the answers to the questions posed in the abstract of this paper as follows:
\begin{itemize}
\item 
We observe that although incremental design can build efficient hierarchies over the iGEM targets, the normalized cost increases over time. This is due to the fact that the amount of reuse from previous years decreases mainly due to the frequent introduction of new sources over time. The small length of the targets in iGEM is also an additional factor for lowering the potential of reuse of the previously constructed parts in iGEM.
\item
The increasing normalized cost causes the Lexis-DAGs to become less deep and to contain shorter nodes on average as time progresses. This is different than Evo-Lexis. In addition, there is a high fraction of very short targets in each year in comparison to Evo-Lexis.
\item
The iGEM Lexis-DAGs  present a bias in reusing specific nodes more often than the other nodes. This biased reuse results in the Lexis-DAGs to take the shape of an hourglass with relatively high H-score values and a stable set of core nodes over time. This observation is consistent with Evo-Lexis.
\item
The core sets over the years remain stable and similar to previous years in iGEM data despite the fact that the set of sources changes significantly and the target sets are diverse each year. Most of the stability is contributed by a small set of central sources and central intermediate nodes that are heavily reused in  iGEM registry over time.
\end{itemize}



\bibliographystyle{splncs04}
{\small

}
\end{document}